\documentclass[conference]{IEEEtran}
\usepackage{url}
\usepackage{graphicx}
\usepackage{algorithm}
\usepackage[noend]{algpseudocode}
\usepackage{subfigure}
\usepackage{amssymb, amsmath, graphicx, charter, latexsym}
\usepackage{amsthm}
\usepackage{setspace}
\usepackage{enumitem}
\usepackage{ragged2e}
\usepackage{times}
\usepackage{mathtools}
\usepackage{tabu}
\usepackage{epstopdf}
\usepackage{multirow}
\usepackage{verbatim}
\usepackage{color}
\usepackage{lipsum}
\usepackage{mdframed}
\usepackage{bm}
\usepackage[cal=cm]{mathalfa}
\usepackage{lipsum}
\usepackage[compress,sort]{cite}
\usepackage{url}

\DeclareMathAlphabet{\mathpzc}{OT1}{pzc}{m}{it}
\DeclarePairedDelimiterX{\bkt}[1]{(}{)}{ #1}
\DeclarePairedDelimiterX{\sbkt}[1]{[}{]}{ #1}
\DeclarePairedDelimiterX{\lbkt}[1]{\{}{\}}{ #1}

\theoremstyle{definition}

\newtheorem{mydefinition}{Definition}
\newcolumntype{x}[1]{>{\raggedright\arraybackslash}p{#1}}
\newcommand{\mathbbm}[1]{\text{\usefont{U}{bbm}{m}{n}#1}}
\def\baselinestretch{1}

\newmdtheoremenv[linewidth=1pt, skipabove=6pt, skipbelow=6pt]{theorem_md}{Theorem}
\newmdtheoremenv[linewidth=1pt, skipabove=6pt, skipbelow=6pt]{observation}{Observation}
\newmdtheoremenv[linewidth=1pt, skipabove=6pt, skipbelow=6pt]{proposition_md}{Proposition}

\IEEEoverridecommandlockouts
\begin{document}

\title{A Semi-Supervised and Inductive Embedding Model for Churn Prediction of Large-Scale Mobile Games}

\author{
	\IEEEauthorblockN{Xi Liu\textsuperscript{*}\thanks{* This work was done during the authors' internship at Samsung Research America}}
	\IEEEauthorblockA{
		\textit{Texas A\&M University}\\
		College Station, TX, USA\\
		xiliu.tamu@gmail.com
	}\and
	\IEEEauthorblockN{Muhe Xie}
	\IEEEauthorblockA{ 
		\textit{Samsung Research America}\\
		Mountain View, CA, USA\\
		muhexie@gmail.com
	}\and
	\IEEEauthorblockN{Xidao Wen\textsuperscript{*}}
	\IEEEauthorblockA{
		\textit{University of Pittsburgh}\\
	    Pittsburgh, PA, USA\\
		xidao.wen@pitt.edu
	}\and
	\IEEEauthorblockN{Rui Chen}
	\IEEEauthorblockA{ 
		\textit{Samsung Research America}\\
		Mountain View, CA, USA\\
		rui.chen1@samsung.com
	}
	\and
	\hspace{15mm} \IEEEauthorblockN{Yong Ge}
	\IEEEauthorblockA{
		\hspace{15mm} \textit{The University of Arizona}\\
	    \hspace{15mm} Tucson, AZ, USA\\
		\hspace{15mm} yongge@email.arizona.edu
	}\and
	\IEEEauthorblockN{Nick Duffield}
	\IEEEauthorblockA{
		\textit{Texas A\&M University}\\
		College Station, TX, USA\\
		duffieldng@tamu.edu
	}\and
	\IEEEauthorblockN{Na Wang}
	\IEEEauthorblockA{
		\textit{Samsung Research America}\\
		Mountain View, CA, USA\\
		na.wang1@samsung.com
	}
}

% <-this % stops a space
% \thanks{Manuscript received}

\maketitle
%\begin{singlespacing}
\begin{abstract}
Mobile gaming has emerged as a promising market with billion-dollar revenues. A variety of mobile game platforms and services have been developed around the world. One critical challenge for these platforms and services is to understand user churn behavior in mobile games. Accurate churn prediction will benefit many stakeholders such as game developers, advertisers, and platform operators. In this paper, we present the first large-scale churn prediction solution for mobile games. In view of the common limitations of the state-of-the-art methods built upon traditional machine learning models, we devise a novel semi-supervised and inductive embedding model that jointly learns the prediction function and the embedding function for user-app relationships. We model these two functions by deep neural networks with a unique edge embedding technique that is able to capture both contextual information and relationship dynamics. We also design a novel attributed random walk technique that takes into consideration both topological adjacency and attribute similarities. To evaluate the performance of our solution, we collect real-world data from the Samsung Game Launcher platform that includes tens of thousands of games and hundreds of millions of user-app interactions. The experimental results with this data demonstrate the superiority of our proposed model against existing state-of-the-art methods.
\end{abstract}
\begin{IEEEkeywords}
Churn prediction, representation learning, graph embedding, inductive learning, semi-supervised learning, mobile apps
\end{IEEEkeywords}
%\end{singlespacing}
%\vspace{-4mm}

\section{Introduction}
\label{section:introduction}

% \textbf{To be changed: 1. generalize the target of the proposed method to app churn prediction, bulk campaign churn prediction, even churn prediction with large numbers of products, not specific to game apps; 
% 2. validate the definition of churn, provide more references for similar churn definition; note the method does not depend on whether it is 14 days or 14 minutes, that could also extend the application scope of it; 3. re-organize the position of contents, now it is problem$\to$data$\to$model, change it to, problem$\to$model$\to$data, put models related part ahead of data and experimental related parts; when talking about problem, not restricted to Samsung game launcher problem, Figure 1 may be put to other place 4. add discussion and validation of experiments;
% 5. add experiments for Korea users; 6. maybe we can also add one part discussion how the model performs with different churn definition, for ex. 7 days, 2 days, etc.
% 6. change confusion matrix to more reasonable measures (to be discussed with Professor Ge. 7. make it for 10 pages full, now is 1 paragraph less than 10 pages; 8. the caption of table is too close to the table.}

With the wide adoption of mobile devices (e.g., smartphones and tablets), mobile gaming has created a billion-dollar market around the globe. According to Newzoo's  Global Games Market Report~\cite{NewZooReport}, mobile games generated \$50.4 billion revenue in 2017. And it is expected that mobile games will generate \$72.3 billion revenue in 2020, accounting for more than half of the overall game market. This increasingly vital market has driven software and hardware providers of mobile devices (e.g., Apple, Google and Samsung) to provide integrated mobile game platforms and services for end users, game developers and other stakeholders. 
%Examples of such platforms and services include Apple App Store and Samsung Game Launcher~\cite{SamsungGameLauncher}.

Within these mobile game platforms and services, one particularly vital task is predicting churn in gaming apps. Churn prediction is a long-standing and important task in many traditional business applications~\cite{zaki2016fallacy}, the objective of which is to predict the likelihood that a user will stop using a service or product. In our problem setting, the aim is to predict the likelihood that a user will stop using a particular game app in the future. Churn prediction is an important problem for the following reasons. First, the churn rate of a mobile game app is an important business metric to measure the success of the game. Successfully estimating the churn probability of all player-game pairs will allow a game platform to better prioritize its resources for operation and management. Second, by predicting individual churn probabilities, game platforms will be able to devise better marketing strategies to improve user retention. Examples include sending push notifications and providing free items in games to users who are likely to churn. Since, as well known, the acquisition cost for new users is much higher than the retention cost for existing users, successful churn prediction could greatly reduce costs for game developers and platform operators. Third, churn prediction provides direct input to determine the right timing of app recommendation for a game platform. The results of this research will enable testing of the hypothesis that a user is more likely to act on an app recommendation when he/she is about to stop playing other games. %on his/her mobile device.

%\begin{figure}[t] 
%\centering
%\includegraphics[width=.45\textwidth]{figures/screenshot.pd%f}
%% \vspace{-4mm}
%\caption{An illustration of the Samsung Game Launcher %platform}
%\label{fig:screenshot}
% \vspace{-4mm}
%\end{figure}

% In this paper, we propose a novel method for the large-scale churn prediction of mobile games by modeling massive user-app interaction data. The user-app interaction data includes detailed information of opens, closes, installs and uninstalls of game apps for each individual user. This data is collected from the Samsung Game Launcher platform \footnote{https://www.samsung.com/au/support/mobile-devices/how-to-use-game-tools/}, which is pre-loaded in most smartphones manufactured by Samsung. The app provides practical toolsets (e.g., power management, alert management, performance mode and game recommendation) to improve user experience in gameplay. 
%Samsung Game Launcher platform, which is the default control center preloaded on most Samsung smartphones. Fig. \ref{fig:screenshot} provides an illustration of the Game Launcher platform. 
%As a major game app platform that is available for all Samsung smartphones, it is very desirable to accurately predict user's churn of mobile game.
% (see Table \ref{table:references} for a detailed comparison)

There have been several previous studies \cite{hadiji2014predicting, runge2014churn, xie2015predicting, perianez2016churn,  tamassia2016predicting,  drachen2016rapid, xie2016predicting, bertens2017games, viljanen2017measuring, viljanen2017playtime,kim2017churn} on mobile game churn prediction by using traditional machine learning models (e.g., logistic regression, random forests, Cox regression). However, we observe several major limitations of these studies. First, they were developed for predicting churn of one or a few mobile games; none is capable of handling the churn prediction of large-scale mobile apps and users. For real-world applications, a solution needs to handle tens of thousands of mobile games and hundred of millions of user-app interactions on a daily basis. Second, user-app interaction data often comes with rich contextual information (e.g., WiFi connection status, screen brightness, and audio volume), which has never been considered in the existing studies. Third, the existing methods rely on handcrafted features that usually cannot scale well in practice. 

To overcome all these limitations, we propose a novel inductive semi-supervised embedding model that \emph{jointly} learns the prediction function and the embedding function for user-game interaction. The user-app interaction data includes detailed information of opens, closes, installs and uninstalls of game apps for each individual user. This data is collected from the Samsung Game Launcher platform\footnote{https://www.samsung.com/au/support/mobile-devices/how-to-use-game-tools/}, which is pre-loaded in most smartphones manufactured by Samsung. We model the interplay between users and games by an attributed bipartite graph and then learn these two functions by deep neural networks with a unique embedding technique that is able to capture both contextual information and dynamic user-game interaction. Our method is fully automatic and can be easily integrated into existing mobile platforms.

\begin{figure}[t] 
\centering
\includegraphics[width=.3\textwidth]{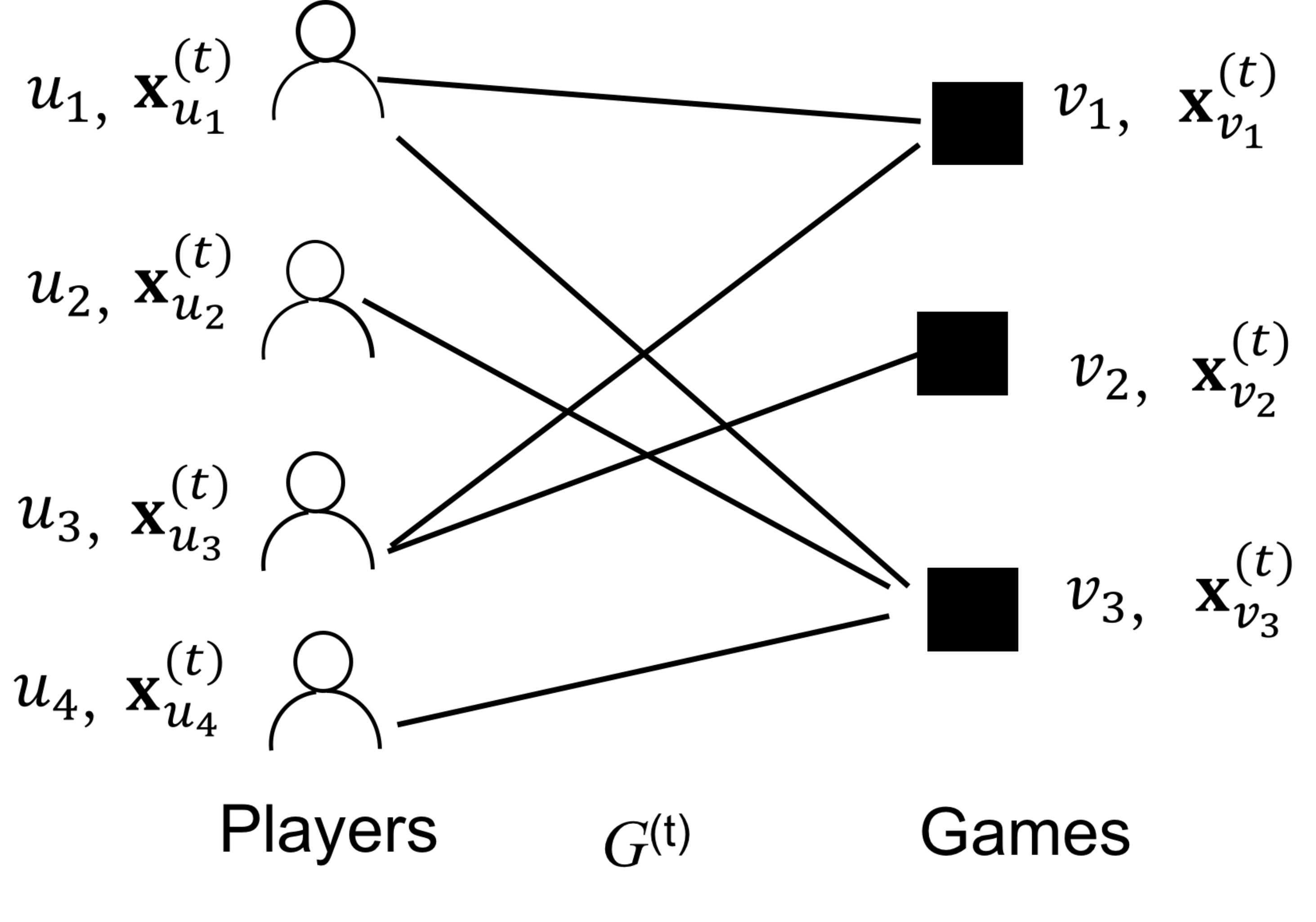}
% \vspace{-4mm}
\caption{An example of an attributed bipartite graph for mobile game churn prediction}
\label{fig:dislink_detection}
\vspace{-4mm}
\end{figure}

\noindent \textbf{Contributions.} Our research contributions are as follows:
%\begin{itemize}
\begin{enumerate}[leftmargin=1pt,itemindent=12pt,itemsep=3pt,parsep=0pt,topsep=0pt,partopsep=0pt]
\item 
%In spite of its important business value, churn prediction of large-scale mobile games has not been adequately studied in the literature. 
To the best of our knowledge, this paper is the first to develop a solution for churn prediction of large-scale mobile games using hundreds of millions of user-app interaction records. This solution has been tested in Game Launcher, one of the largest commercial mobile gaming platforms. Although the paper mainly applies the proposed solution in mobile game churn prediction, the solution is also applicable to churn prediction in other contexts.

\item We propose a novel semi-supervised and inductive model based on embedding. Our model can capture the dynamics between users and mobile games based on the introduced temporal loss in the formulated objective function. The model is able to embed new users or games not used in training. This is critical for mobile game churn prediction because new games and users continually enter the market. 

\item We develop an attributed random walk technique that enables us to sample the contexts of edges in an attributed bipartite graph and that takes into account both topological adjacency and attribute similarities.

\item We conduct a comprehensive experimental evaluation with large-scale real-world data collected from Samsung Game Launcher. The experimental results demonstrate that our model outperforms all state-of-the-art methods with respect to different evaluation metrics for prediction.
\end{enumerate}
%\end{itemize}

% Paper organization
%The rest of the paper is organized as follows. Section~\ref{sec:problem} formulates the mobile game churn prediction problem. Section~\ref{section:methods} discusses our solution in detail. Section~\ref{section:experimental_evaluation} presents our experimental results on large-scale real-world data. Section~\ref{section:related_work} reviews the related literature. Finally, Section~\ref{section:conclusion} concludes our work.

\section{Problem Formulation}
\label{sec:problem}
In the context of mobile games, \emph{churn} is defined as a player stopping using a game within a given period (i.e., there is no app usage in the period). The duration $T$ of the period may vary from application to application depending on different business goals. $T=14$ days and $T=30$ days are some typical settings used in industry \cite{hadiji2014predicting,runge2014churn,drachen2016rapid}. In this paper, we consider the generic game churn prediction problem without assuming any particular value of $T$. We note that uninstall is different from churn. Regarding only uninstall as churn would be problematic since there may be a large time gap between cessation of playing and uninstall, if any. 

%Instead, we use the customary definition of ``churn'' used in most papers, i.e., no in-app activity for $T$ consecutive time units.
%(the length of one time unit can be adjusted for specific application scenarios). 
%This definition is flexible and in most cases is a stronger indicator than uninstallation to detect interests lost. 
%The choice of $T$ depends on specific prediction needs. Smaller $T$ encompasses more cases and is hence more sensitive; conversely larger $T$ is more selective for longer term app non-use.  
%The method we proposed in this paper is not restricted to the choice of $T$. 

The relationship between players and games can be represented by an attributed bipartite graph as illustrated in Fig.~\ref{fig:dislink_detection}, whose two parts correspond to players and games. In the sequel, we use the terms \emph{player} and \emph{user}, and \emph{game} and \emph{app} interchangeably. Let $\mathcal{G}^{(t)}$ be the attributed bipartite graph at time $t$, the vertex set $U^{(t)}$ denote the set of users and the vertex set $ V^{(t)}$ denote the set of games. A player is represented by a node $u\in U^{(t)}$ and a game is represented by a node $v \in V^{(t)}$. Each user $u$ is associated with a feature vector $\mathbf{x}_{u}^{(t)}\in \mathcal{R}^{n_{u}}$, where $n_u$ is the size of $\mathbf{x}_{u}^{(t)}$; each game $v$ is associated with a feature vector $\mathbf{x}_{v}^{(t)}\in \mathcal{R}^{n_{v}}$, where $n_{v}$ is the size of $\mathbf{x}_{v}^{(t)}$. There is an edge between nodes $u$ and $v$ in $\mathcal{G}^{(t)}$ if player $u$ has played $v$ in the time window $[t+1, t+T]$. The set of edges is denoted by $E^{(t)}$. 
%An attributed bipartite graph is illustrated in Fig.\ref{fig:dislink_detection}.

\begin{table}[t]
\centering
\caption{Notations and Definitions}
{\small
\begin{tabular}{|p{1.3
cm}<{\centering}|p{6.5cm}<{\centering}|}
\hline
Notations & Descriptions or Definitions 
\\ \hline
$\mathcal{G}^{(t)}$ & Attributed graph at time $t$
\\ \hline
$\mathcal{G}^{(t+1)}$ & Attributed graph at time $t+1$
\\ \hline
$\mathcal{H}^{(t)}$ & Historical attributed graphs $\{\mathcal{G}_{i}\}_{i = 0}^{i=t}$
\\ \hline
$U^{(t)}$ & Set of all player nodes in $\mathcal{G}^{(t)}$
\\ \hline
$V^{(t)}$ & Set of all game nodes in $\mathcal{G}^{(t)}$
\\ \hline
$E^{(t)}$ & Set of all edges in $\mathcal{G}^{(t)}$
\\ \hline
$e_{uv}^{(t)}$ & Indicator of the existence of edge $(u,v)$ in $\mathcal{G}^{(t)}$
\\ \hline
$\mathbf{x}_{u}^{(t)}$ & Feature vector of user $u\in U^{(t)}$
\\ \hline
$\mathbf{x}_{v}^{(t)}$ & Feature vector of game $v \in V^{(t)}$
\\ \hline
$\mathbf{z}_{uv}^{(t)}$ & Aggregated feature vector of edge $(u,v) \in E^{(t)}$
\\ \hline
$n_u$, $n_v$ & Numbers of attributes in $\mathbf{x}_{u}^{(t)}$ and $\mathbf{x}_{v}^{(t)}$, resp.
\\ \hline
$d$ & Number of attributes in $\mathbf{z}_{uv}^{(t)}$
\\ \hline
$m$ & Embedding dimension
\\ \hline
$g$ & The edge embedding function $g:\mathcal{R}^{d} \rightarrow \mathcal{R}^{m}$
\\ \hline
$f$ & The churn prediction function $f: \mathcal{R}^{m} \rightarrow [0, 1]$
\\ \hline
$l_{n}$ & Number of prediction hidden layers in Part III
\\ \hline
$l_{p}$ & Number of embedding hidden layers in Part I
\\ \hline
\end{tabular}
}
\label{table:notations}
\vspace{-4mm}
\end{table}

\begin{figure*}[t]
\centering
\includegraphics[width=1\textwidth, height=.425\textwidth]{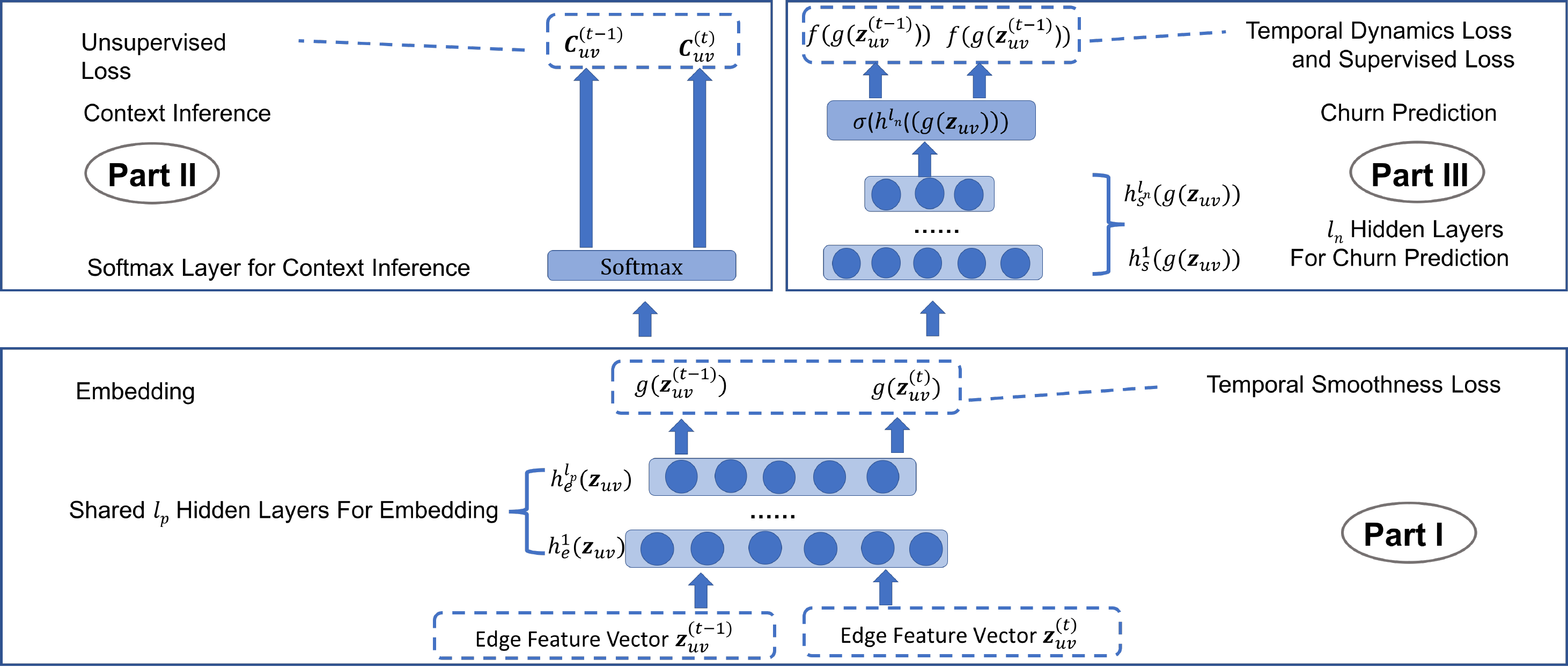} % Give the name of the image here.
\vspace{-5mm}
\caption{Deep neural network architecture of the inductive semi-supervised embedding model in training}
\label{fig:neural_net_archi}
\vspace{-2mm}
\end{figure*}

Now we are ready to define the mobile game churn prediction problem as follows\footnote{This problem definition is indeed generic to be applied to churn prediction in other contexts, for example, general app churn prediction.}.
\begin{mydefinition} [\emph{Mobile game churn prediction}]
Consider a collection of attributed bipartite graphs observed from time $t_0$ to time $t$ ($t > t_0$), which is denoted by $\mathcal{H}^{(t)} = \{\mathcal{G}^{(i)}\}_{i = t_0}^{i=t}$. Let $e_{uv}^{(i)}$ be the indicator of the existence of edge $(u, v)$ in $\mathcal{G}^{(i)}$. For any edge $(u, v)$ in $\mathcal{G}^{(t)}$, predict the probability $Pr(e_{uv}^{(t+1)} = 0 | e_{uv}^{(t)} = 1, \mathcal{H}^{(t)})$, that is, the probability that $(u, v)$ disappears in $\mathcal{G}^{(t+1)}$.
\end{mydefinition}

The notations and their descriptions are listed in Table～\ref{table:notations}.

%$\mathbf{x}_{u}^{(t)}$ and $\mathbf{x}_{v}^{(t)}$ are constructed from historical user-app interaction data. That is, suppose the observation starts from $t=0$, historical attributed bipartite graphs $\mathcal{H}^{(t)} = \{\mathcal{G}_{i}\}_{i = 0}^{i=t}$. There is an edge between node $u$ and $v$ in $\mathcal{G}^{(t)}$ if player $u$ has played game $v$ during period $t+1 \sim t+T$. In the discussion below, the attributed graph refers both the structure of the graph and the attributes of nodes.

% \noindent As shown in Fig.\ref{fig:dislink_detection}, suppose $u_{1}$ churns $v_{3}$ at time $t+1$, then one should observe that $(u_{1},v_{3})$ exists in $\mathcal{G}^{(t)}$ but not exists in $\mathcal{G}^{(t+1)}$. The problem can then be formulated as following way: given attributed graph $\mathcal{G}^{t}$ for some time $t$, for each edge $(u,v)$ in $\mathcal{G}^{(t)}$, learn the probability that $(u,v)$ disappears in $\mathcal{G}^{t+1}$. Since player and game feature vectors are constructed from $\mathcal{H}^{(t)}$, that can be expressed as

\section{Methods}\label{section:methods}
% Graph embedding automates the whole process of feature engineering, which frees model from human bottleneck introduced by handcrafted features and is able to utilize the full richness of the data. However, most existing works are node-centric embedding, not edge-centric. Meanwhile, they belong to transductive framework, which cannot produce embedding for new entities. 
% \vspace{-4mm}
 % \vspace{-2mm}

\subsection{Overview of Our Solution}

Most existing works~\cite{hadiji2014predicting, runge2014churn, xie2015predicting, perianez2016churn,  tamassia2016predicting,  drachen2016rapid, xie2016predicting, bertens2017games, viljanen2017measuring, viljanen2017playtime,kim2017churn} rely on traditional machine learning models to solve the churn prediction problem, which suffer from several major limitations identified in Section~\ref{section:introduction}. In view of the recent progress in deep learning and graph embedding, a natural promising direction is to adopt graph embedding frameworks for churn prediction. However, we face several key technical challenges that have not been addressed by existing research: (1) All existing methods are transductive, and thus cannot produce embeddings for new player-game pairs. (2) Existing methods are either purely supervised or unsupervised, and thus do not take full advantage of relevance between embedding and a task. (3) Existing methods are node-centric, and thus are not directly applicable to edge related tasks. (4) Existing methods only handle a static graph and do not incorporate graph dynamics in embedding. In the mobile game industry, players and games, however, change very quickly. There are new players, new games, and new player-game relationships every day. 

In addressing these challenges, we propose a novel inductive semi-supervised embedding model in dynamic graphs that jointly learns the prediction function $f$ and the embedding function $g$. The prediction function $f$ and the embedding function $g$ are learned by deep neural networks (DNN). The architecture of the proposed DNN is presented in Fig. \ref{fig:neural_net_archi}. The DNN consists of three parts. Part I is responsible for producing embedding feature vectors $g(\mathbf{z}_{uv}^{(t)})$ from raw edge feature vectors $\mathbf{z}_{uv}^{(t)} \in \mathcal{R}^{d}$, where $d$ is the size of raw edge feature vectors. To learn the probability of churn, we need to construct a feature vector for each $(u,v)$ with $e_{uv}^{(t)} = 1$. However, it is impractical to calculate features for all possible edges which may appear in the prediction period because the number of possible edges is large, which is $\mathcal{O}(|U^{(t)}|\cdot |V^{(t)}|)$. Instead, we construct the feature vector $\mathbf{z}_{uv}^{(t)}$ of $(u,v)$ from attribute-wise cosine similarity aggregation of $\mathbf{x}_{u}(t)$ and $\mathbf{x}_{v}(t)$. As such, all $\mathbf{z}_{uv}^{(t)}$ can readily computed from $|U^{(t)}| +|V^{(t)}|$ node features, where $|\cdot|$ denotes the cardinality of a set. 

%$\mathbf{z}_{uv}^{(t)} $ 
%The raw edge features are constructed by attribute-wise aggregating user and game feature vectors $\mathbf{x}_{u}^{(t)}$ and $\mathbf{x}_{v}^{(t)}$ using cosine similarity. 

Part II is in charge of inferring contexts from embedding feature vectors. Here the context of an edge refers to the edges that are similar to and co-occur with the edge under some graph sampling strategy, for example, random walk. Part I and Part II make up the \emph{unsupervised} component of our model. They are jointly trained by minimizing the error due to incorrect context inference and inconsistency with temporal smoothness (see Section~\ref{subsection:temporal_loss} for an explanation of temporal smoothness). Part I and Part II are trained in an inductive and edge-centric way. In contrast to transductive node embedding that learns a distinct embedding vector for each node, our idea is to learn an embedding function that generalizes to any \emph{unseen} edges as long as their feature vector is available. Since producing contexts of edges has not be previously studied, we propose a novel attributed random walk to sample similar edges as contexts (see Section~\ref{subsection:attributed_random_walk} for details). 

Part III fulfills the supervised churn prediction task from embedding feature vectors. Part III forms the \emph{supervised} component of the proposed model, which is trained by minimizing the error of incorrect churn predictions. The supervised component and unsupervised component are simultaneously trained in a single objective function. Traditional unsupervised embedding techniques are not designed in a task-specific way and hence are not able to incorporate task-specific information to improve performance. In contrast, in our model Part III and Part II share the common hidden layers in Part I, and therefore they are latently coupled with each other. This helps the embedding align with the supervised prediction task. 

Part I and Part III both consider graph dynamics in training. Part I handles graph dynamics by requiring the embeddings of the same edge at two consecutive timestamps to stay close. Part III handles graph dynamics by requiring the churn probabilities of the same edge at two consecutive timestamps to follow a decaying pattern.

The objective function of our model is composed of four parts:
\begin{align}\label{equ:objective}
    L := L_{S} + \alpha L_{U} + \beta  L_{T} + \gamma L_{R}.
\end{align}
$L_{S}$ denotes the supervised loss due to incorrect predictions and will be discussed in Section~\ref{subsection:supervised_loss}. $L_{U}$ denotes unsupervised loss, which comes from failures of context inference and will be addressed in Section~\ref{subsection:unsupervised_loss}. $L_{T}$ is the temporal loss that consists of two parts: temporal smoothness and temporal dynamics, and will be explained in Section~\ref{subsection:temporal_loss}. $L_{R}$ presented in Section~\ref{subsection:reg_loss} is the regularization term, and ($\alpha$, $\beta$, $\gamma$) are trade-off weights.

\subsection{Static Loss Functions}
\label{subsection:loss_functions}
%----------------------------------------------
\subsubsection{Supervised Loss Function $L_{S}$}\label{subsection:supervised_loss}
The supervised loss function $L_{S}$ is designed for Part III. Let $h_{s}^{k}(g(\mathbf{z}_{uv}^{(t)})) = \phi(W_{s}^{k}h_{s}^{k-1}(g(\mathbf{z}_{uv}^{(t)})) + b^{k})$ be the $k$-th hidden layer for churn prediction (referred to as \emph{prediction hidden layer} in the sequel), where $W^{k}$ and $b^{k}$ are the weights and biases in the $k$-th prediction hidden layer, and $\phi(\cdot)$ is a non-linear activation function. We model the churn prediction function $f$ by $l_{n}$ such layers in Part III. Then the prediction output layer can be represented by:
\begin{align}\label{equ:supervised_loss}
    f(g(\mathbf{z}_{uv}^{(t)})) &:=Pr(e_{uv}^{(t+1)} = 0 | e_{uv}^{(t)} = 1, \mathcal{H}^{(t)}) \\ \nonumber &:=\sigma(h_{s}^{l_{n}}(g(\mathbf{z}_{uv}^{(t)}))) \\ \nonumber
    &:= \frac{\exp{\big(h_{s}^{l_{n}}(g(\mathbf{z}_{uv}^{(t)}))^{T}w_{s}}\big)}{1+\exp{\big(h_{s}^{l_{n}}(g(\mathbf{z}_{uv}^{(t)}))^{T}w_{s}}\big)}
\end{align}
%\begin{align}\label{equ:supervised_loss}
%    f(g(\mathbf{z}_{uv}^{(t)})) &:=Pr(e_{uv}^{(t+1)} = 0 | e_{uv}^{(t)} = 1, \mathcal{H}^{(t)}) \\ \nonumber &:=\sigma(h_{s}^{l_{n}}(g(\mathbf{z}_{uv}^{(t)}))) \\ \nonumber
%    &:= \frac{\exp{\big(h_{s}^{l_{n}}(g(\mathbf{z}_{uv}^{(t)}))^{T}w_{s}}\big)}{1+\exp{\big(h_{s}^{l_{n}}(g(\mathbf{z}_{uv}^{(t)}))^{T}w_{s}}\big)},\\ \nonumber
%    h_{s}^{k}(\mathbf{z}_{uv}^{(t)}) &:= \phi(W_{s}^{k}h_{s}^{k-1}(g(\mathbf{z}_{uv}^{(t)})) + b_{s}^{k}), \nonumber
%\end{align}
where $\sigma(\cdot)$ is the sigmoid function and $w_{s}$ is the sigmoid weights vector that combines the output from the last hidden layer to predict churn. Now we can define the supervised loss $L_{S}$ as follows:
\begin{align}
    L_{S} =& \frac{1}{L}\sum_{i = 0}^{t-1} \sum_{(u,v) \in E^{(i)}} \delta_{uv}^{(i+1)}\Big(1-e_{uv}^{(i+1)} - \\\nonumber
    &\frac{\exp{\big(h_{s}^{l_{n}}(g(\mathbf{z}_{uv}^{(i)}))^{T}w_{s}}\big)}{1+\exp{\big(h_{s}^{l_{n}}(g(\mathbf{z}_{uv}^{(i)}))^{T}w_{s}}\big)}\Big)^{2},
\end{align}
 where $L$ is the number of training examples and $\delta_{uv}^{(i+1)}$ is a censoring indicator and will be discussed in Section \ref{subsection:temporal_loss}.
% Denote by the $g(\mathbf{z}_{uv}(t))$ the embedding for edge $(u,v)$ in $G_{t}$, and the  the predicted churn probability for edge $(u,v)$ at time $t+1$.
% \begin{align}
%     L_{S} := \sum_{i = 0}^{t-1} \sum_{(u,v) \in E^{(i)}} \delta_{uv}^{(i+1)}\ell(e_{uv}^{(i+1)}, f(g(\mathbf{z}_{uv}^{(i)})),
% \end{align}
% where $\ell$ is the loss function.
%----------------------------------------------
\subsubsection{Unsupervised Loss Function $L_{U}$}\label{subsection:unsupervised_loss}
The unsupervised loss function $L_{U}$ is devised for Part II, which guides to embed the handcrafted features $\mathbf{z}_{uv}^{(t)}\in \mathcal{R}^{d}$ into a latent space $g(\mathbf{z}_{uv}^{(t)})\in \mathcal{R}^{m}$, where $m$ is the size of the latent space. Denote the $k$-th hidden layer for embedding (referred to as \emph{embedding hidden layer} in the sequel) by $h_{e}^{k}(\mathbf{z}_{uv}^{(t)}) = \phi(W_{e}^{k}h_{e}^{k-1}(\mathbf{z}_{uv}^{(t)}) + b_{e}^{k})$, where $W_{e}^{k}$ and $b_{e}^{k}$ are the weights and biases in the $k$-th embedding hidden layer. We use the $l_{p}$ layers in Part I to represent the embedding function $g$, and the embedding output layer can be represented by:
\begin{align}\label{equ:unsupervised_loss}
    &g(\mathbf{z}_{uv}^{(t)}) := h_{e}^{l_{p}}(\mathbf{z}_{uv}^{(t)}).
\end{align}

%\\
%    &h_{e}^{k}(\mathbf{z}_{uv}^{(t)}) := \phi(W_{e}^{k}h_{e}^{k-1}(\mathbf{z}_{uv}^{(t)}) + b_{e}^{k}). \nonumber
   
We can define the unsupervised loss function as follows:
{\small
\begin{align}
    L_{U} = -\sum_{i=t_0}^{t}\sum_{(u,v) \in E^{(i)}} \sum_{(u',v')\in C_{uv}^{(i)}}\delta_{uv}^{(i)} \log\Big(\Pr\big((u',v')|g(\mathbf{z}_{uv}^{(i)}\big)\Big),
\end{align}
}%
where $C_{uv}^{(i)}$ denotes the context (i.e., contextual edges) of $(u,v)$ in $\mathcal{G}^{(i)}$. The contextual edges $C_{uv}^{(i)}$ are obtained by attributed random walk on the bipartite graph, which will be discussed in Section~\ref{subsection:attributed_random_walk}.

The likelihood of having a contextual edge $(u', v')$ of $(u,v)$ conditional on the embedding of $(u,v)$ is:
\begin{align}
    Pr\big((u',v')|g(\mathbf{z}_{uv}^{(i)})\big) = \frac{\exp\Big( g(\mathbf{z}_{uv}^{(i)})^{T}w_{u'v'}\Big)}{{\sum_{(u^{*},v^{*})}\exp\Big(g(\mathbf{z}_{uv}^{(i)})^{T}w_{u^{*}v^{*}}\Big)}},
\end{align}
where $w_{u^{*}v^{*}}$ and $w_{u'v'}$ are the vectors of weights for edges $(u^{*},v^{*})$ and $(u', v')$ in the softmax layer, respectively. The denominator is computed by negative sampling \cite{liang2017seano}. Although the embedding function is learned by training a context inference task, it is still considered as ``unsupervised'' because the contexts are calculated by sampling on the attributed graph, which is independent of any supervised learning task~\cite{yang2016revisiting}.

% Part I and II in Fig. \ref{fig:neural_net_archi} correspond to the unsupervised components responsible for embedding. 

The objective function in Equation (\ref{equ:objective}) contains both the supervised loss function and the unsupervised loss function. Thus the embedding hidden layers are jointly trained with prediction hidden layers. Compared to traditional embedding methods, the semi-supervised approach makes the embedding more suitable for the prediction task. 
%This defines the proposed model follows a semi-supervised framework. 

Note that the target of the embedding process is to learn a mapping function from the feature space to the embedding space, instead of directly learning the embeddings. Thus, the input for the embedding hidden layers only contains attributes. A new edge can be embedded for churn prediction as long as we can observe its attributes. This indicates that the proposed approach is inductive. It is able to produce the embedding for an edge that is not used in training.

\subsubsection{Regularization Loss $L_{R}$}\label{subsection:reg_loss}
Regularization loss is introduced mainly to avoid over-fitting. The weights for regularization consists of $\{\{W^{i}_{e}, b_{e}^{i}\}_{i=1}^{l_{p}}, \{W^{i}_{s}, b_{s}^{i}\}_{i=1}^{l_{n}}, w_{s}\}$. Therefore the regularization part can be expressed as:
\begin{eqnarray}
    L_{R} &:=& \lambda_{0}\sum_{i = 1} ^{l_{p}}\| W^{i}_{e} \|^{2}_{2} +\lambda_{1}\sum_{i = 1} ^{l_{p}}\| b^{i}_{e}\|^{2}_{2} +  \lambda_{2}\sum_{i = 1} ^{l_{n}}\| W^{i}_{s} \|^{2}_{2}  \nonumber \\
     &&+\lambda_{3}\sum_{i = 1} ^{l_{n}}\| b_{s}^{k}\|^{2}_{2}  + \lambda_{4}\| w_{s}\|^{2}_{2},
\end{eqnarray}
where $\{\lambda_{i}\}_{i=0}^{4}$ are trade-off weights on different regularization terms.

%\vspace{-1mm}
\subsection{Temporal Loss Function}
\label{subsection:temporal_loss}
Temporal loss refers to the loss related to graph dynamics. Different from existing works~\cite{liang2017seano,yang2016revisiting,hamilton2017inductive}, the proposed embedding model takes into account graph dynamics. Understanding the temporal dynamics of attributed bipartite graphs is crucial to precisely model the churn behavior. We make several key observations of the temporal dynamics, which help to achieve good prediction performance in practical settings. 

% The temporal loss function consists of two parts corresponding to two types of dynamics. The first term corresponds to the temporal smoothness, i.e., in most cases, the embedding of the same player-game pair that exists in two consecutive days does not change much. The second term corresponds to the temporal dynamics, i.e., the churn probability for a specific player-game pair increases as the days of play increases . 
\begin{observation}
\label{obs1}
\vspace{-2mm}
For a given user-game play relationship, the longer the relationship exists, the more likely the user is to churn the game.
\end{observation}

\begin{figure}
\centering
\includegraphics[width=.4\textwidth, height=.24\textwidth]{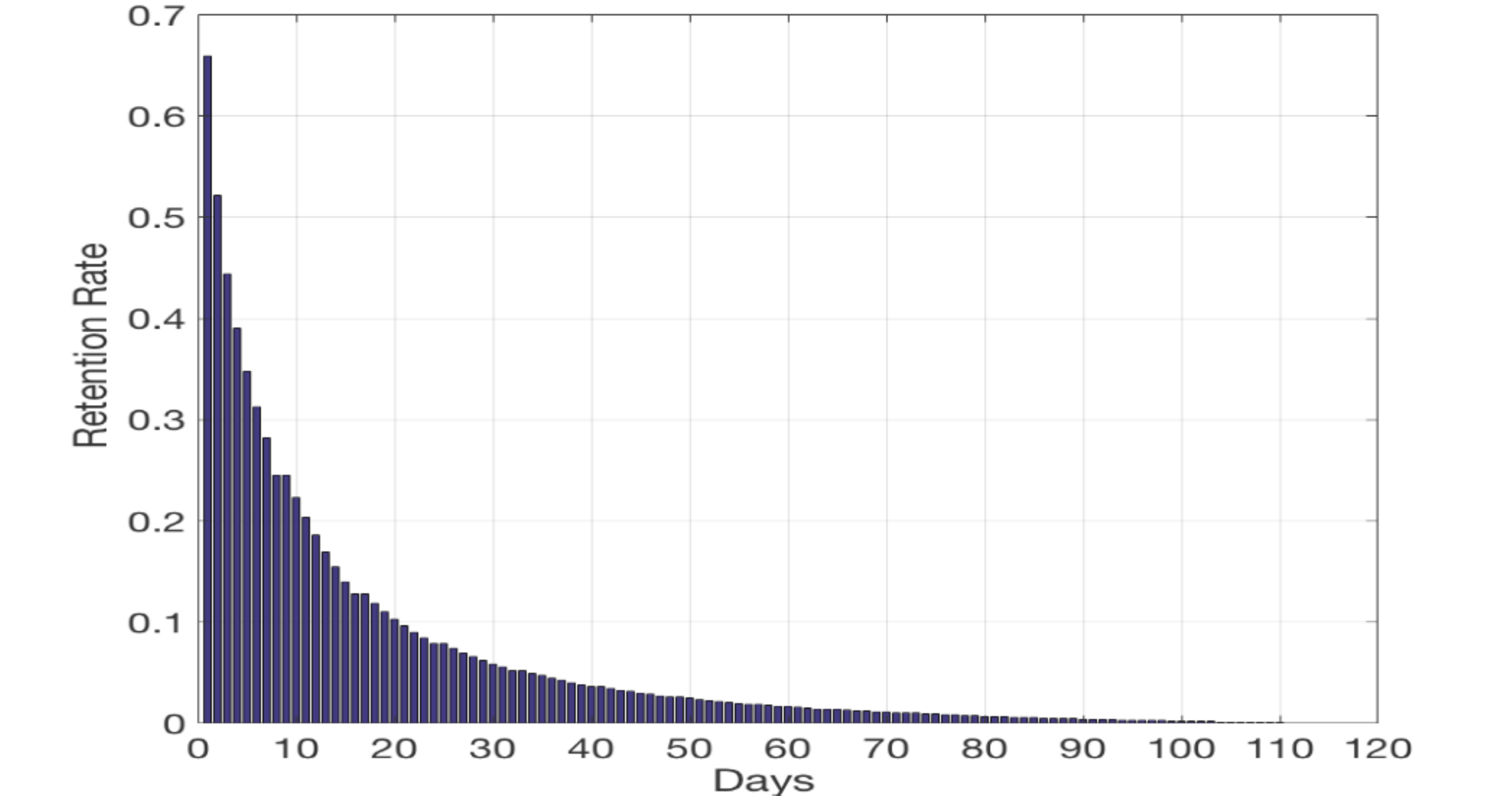}
%\vspace{-2mm}
\caption{Average mobile game retention rate as a function of time}
%\vspace{-4mm}
\label{fig:life_value}
\vspace{-1mm}
\end{figure}

This is because the content of a mobile game is usually somewhat fixed. Players can easily lose interests after going through all contents and passing all levels in the game, let alone many players churn before passing all levels. With more days of play, their initial interests in the game are gradually effacing. Indeed, $71\%$ of all mobile app users churn within 90 days~\cite{AppChurn}. The churn rate of mobile games is even higher. We plot the average retention rate of mobile games as a function of time, based on Game Launcher data in Fig.~\ref{fig:life_value}. It shows that 95\% of user-game play relationships end after 40 days, which well justifies our observation. We formally state Observation~\ref{obs1} below.
\begin{align}\label{equ:temporal_dynamics}
    f(g(\mathbf{z}_{uv}^{(i)})) \lessapprox f(g(\mathbf{z}_{uv}^{(i+1)})) \ \forall 0\leq i \leq t-1, (u,v) \in D^{(i)},
\end{align}
where $D^{(i)} = \{(u,v): (u,v) \in E^{(i)}\cap E^{(i+1)} \}$ and $\lessapprox$ represents ``almost always smaller than''. We refer to this observation as \emph{temporal dynamics} in the following discussion. 

% \vspace{-4mm}

The second observation we make is stated in Observation~\ref{obs2}.
\begin{observation}
\label{obs2}
\vspace{-2mm}
For a given user-game play relationship denoted by an edge in the attributed bipartite graph, its context usually evolves slowly at two consecutive timestamps.
\end{observation}

This observation is also derived from the real-world data. Due to space limitation, we omit the figure here. It follows that the topology and attribute values of the attributed bipartite graph mostly evolve smoothly at two consecutive timestamps, resulting in similar contexts for a given edge at consecutive timestamps. Therefore, its embeddings at these two timestamps should also be close, that is,
\begin{align}\label{equ:temporal_smoothness}
    g(\mathbf{z}_{uv}^{(i+1)}) \approxeq g(\mathbf{z}_{uv}^{(i)}) \ \forall 0\leq i \leq t-1, (u,v) \in D^{(i)},
\end{align}
where $\approxeq$ represents ``almost always equal to''. We refer to this observation as \emph{temporal smoothness} in the later discussion.

\begin{figure}
\centering
\includegraphics[width=.46\textwidth]{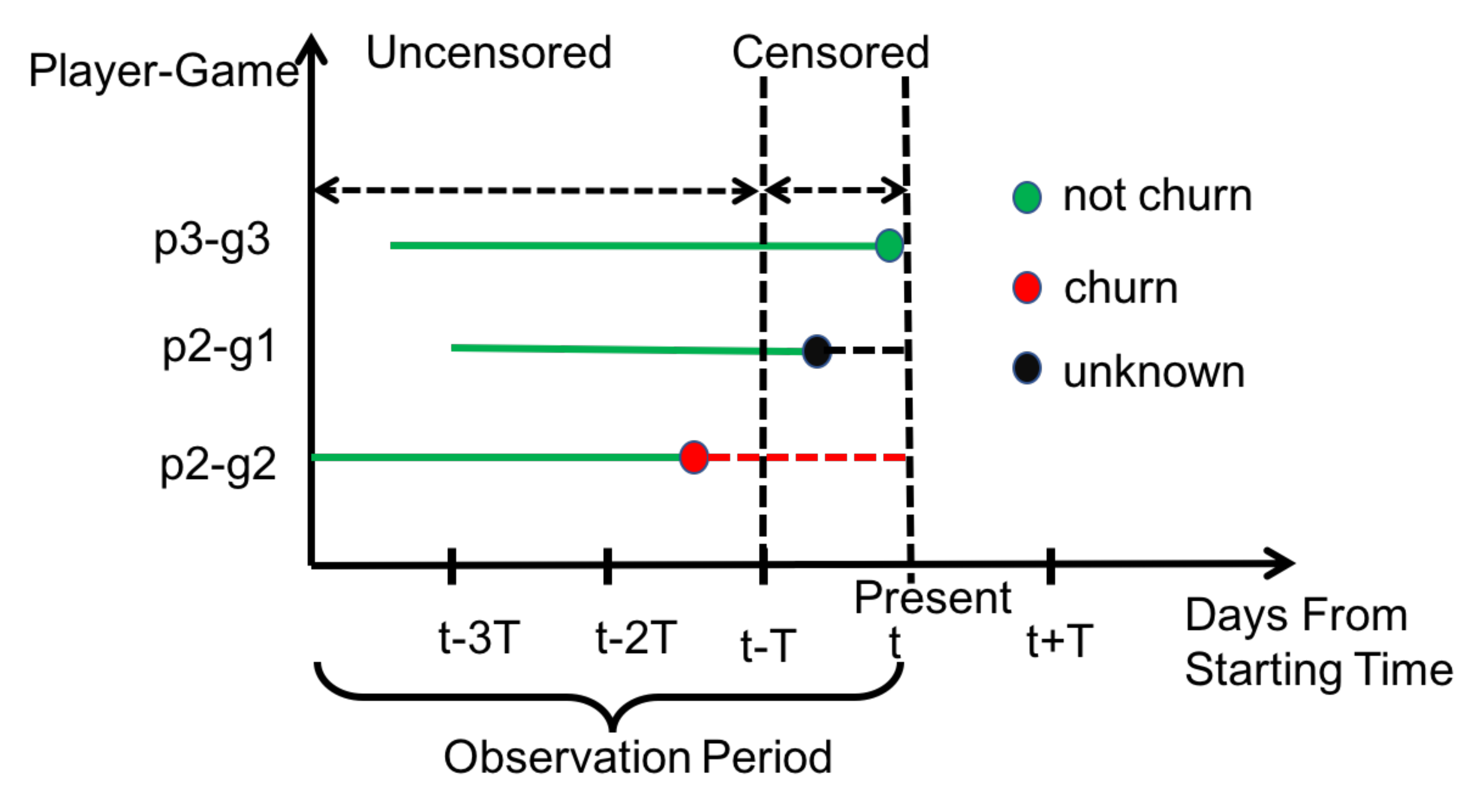}
\vspace{-2mm}
\caption{An illustration of censored data by three randomly sampled player-game pairs}
\label{fig:censored}
\vspace{-3mm}
\end{figure}

The final observation is:
\begin{observation}
\label{obs3}
\vspace{-2mm}
By definition, churn in nature introduces right censoring to the training dataset.
\end{observation}
% \begin{lemma_md}
% For any strictly feasible vector $\bm{q}\in \mathcal{Q}^{*}$, there exists a stationary randomized policy that fulfills $\bm{q}$ based on a probability distribution that only depends on the number of packets to be delivered and the number of time slots remaining in the current interval.
% % \vspace{1mm}
% \end{lemma_md}
% \vspace{-4mm}
The problem of right censoring has been widely studied~\cite{wu1988estimation} and is illustrated in Fig.~\ref{fig:censored}. %illustrates an example of censored training instances.
The observation period refers to some time duration in history. Suppose we are at time $t$ and the observation period is from time $t_0$ to time $t$. Data for training and testing all come from the observation period. Since the label of a player-game pair at a specific timestamp is determined by their interaction in the next $T$ time duration, the labels of some player-game pairs in the last $T$ duration could be \emph{unknown}. For instance, the last observation of the pair of player 2 and game 1 (denoted by p2-g1) was in the last $T$ time duration, and therefore the labels after that time are unknown. This is known as right censoring. In contrast, the pair p3-g3 has play records every day in the last $T$ days, and it is not censored during the observation period.

Considering the existence of censored instances, we introduce a binary indicator $\delta_{uv}^{(i)}$ to indicate whether an edge $(u, v)$ is censored at timestamp $i$. $\delta_{uv}^{(i)}=0$ if $(u, v)$ is censored; $\delta_{uv}^{(i)} = 1$ otherwise. Then Inequality～(\ref{equ:temporal_dynamics}) needs to be updated by
\begin{align}\label{equ:censor}
    &f(g(\mathbf{z}_{uv}^{(i)})) \lessapprox f(g(\mathbf{z}_{uv}^{(i+1)})) \ \forall \ 0\leq i \leq t_{uv}-1, (u,v) \in D^{(i)} \\ 
    &f(g(\mathbf{z}_{uv}^{(t_{uv})}) \lessapprox f(g(\mathbf{z}_{uv}^{(i)})) \ \forall \ t_{uv}\leq i \leq t, (u,v) \in E^{(t_{uv})}, \nonumber
\end{align}
where $t_{uv}$ denotes the timestamp when the edge $(u,v)$ was observed, i.e., $t_{uv} = \max\{i: \delta_{uv}^{(i)} = 1\}$. The update reflects the fact that after timestamp $t_{uv}$ the label of $(u,v)$ becomes unknown. Since the existence of edge $(u, v)$ after $t_{uv}$ is unknown, it is more reasonable to just require that Inequality~(\ref{equ:censor}) holds \emph{pairwise} between time point $t_{uv}$ and all time points after $t_{uv}$. Therefore the temporal loss can be expressed as:
\begin{align}\label{equ:temporal_loss}
    L_{T} :=& \sum_{i = t_0}^{t-1} \sum_{(u,v) \in D^{(i)}}  \Big\{\|g(\mathbf{z}_{uv}^{(i+1)}) - g(\mathbf{z}_{uv}^{(i)})\|_{2} + \\ \nonumber
    &\big[\mathbbm{1}(\delta_{uv}^{(i+1)}=1)f(g(\mathbf{z}_{uv}^{(i)})) +  \\\nonumber
    &\mathbbm{1}(\delta_{uv}^{(i+1)}=0)f(g(\mathbf{z}_{uv}^{(t_{uv})})) - f(g(\mathbf{z}_{uv}^{(i+1)}))\big]_{+}\Big\},
\end{align}
where $\mathbbm{1}(A)$ denotes the indicator function for event $A$ and $[x]_{+} = \max\{x, 0\}$. The first term corresponds to the temporal smoothness and the second term corresponds to the temporal dynamics. Taking Equations (\ref{equ:supervised_loss}) and (\ref{equ:unsupervised_loss}) into Equation (\ref{equ:temporal_loss}), we have the temporal loss $L_{T}$ as follows.
\begin{align}
    L_{T} :=& \sum_{i = 0}^{t-1} \sum_{(u,v) \in D^{(i)}}  \Big\{\|h_{e}^{l_{p}}(\mathbf{z}_{uv}^{(i+1)}) - h_{e}^{l_{p}}(\mathbf{z}_{uv}^{(i)})\|_{2} + \\ \nonumber
    &\big[\mathbbm{1}(\delta_{uv}^{(i+1)}=1)\frac{\exp{\big(h_{s}^{l_{n}}(g(\mathbf{z}_{uv}^{(i)}))^{T}w_{s}}\big)}{1+\exp{\big(h_{s}^{l_{n}}(g(\mathbf{z}_{uv}^{(i)}))^{T}w_{s}}\big)} +  \\\nonumber
    &\mathbbm{1}(\delta_{uv}(i+1)=0)\frac{\exp{\big(h_{s}^{l_{n}}(g(\mathbf{z}_{uv}^{(t_{uv})}))^{T}w_{s}}\big)}{1+\exp{\big(h_{s}^{l_{n}}(g(\mathbf{z}_{uv}^{(t_{uv})}))^{T}w_{s}}\big)} - \\ \nonumber &\frac{\exp{\big(h_{s}^{l_{n}}(g(\mathbf{z}_{uv}^{(i+1)}))^{T}w_{s}}\big)}{1+\exp{\big(h_{s}^{l_{n}}(g(\mathbf{z}_{uv}^{(i+1)}))^{T}w_{s}}\big)}\big]_{+}\Big\}.
\end{align}
%\vspace{1mm}

% larger than that in day $t_{uv}$, but may be wrong to still constrain pairwise  relationship for days after $t_{uv}$. Similarly, for Equation (\ref{equ:temporal_smoothness}), $E^{(i)}$ needs to be changed to $E^{(i)}=\{(u,v): e_{uv}^{(i)} = 1 \vee \delta_{uv}^{(i)} = 0\}$.

\subsection{Context Generation by Attributed Random Walk}
\label{subsection:attributed_random_walk}
All above discussion assumes the availability of contexts of edges. There have been a series of node-centric research on graph embedding proposing to apply random walk to sample contexts~\cite{grover2016node2vec}. These methods are normally topology-based. In our problem setting, our goal is to embed an \emph{edge}, not a node. In this case, a simple topology-based random walk may return two adjacent edges having the same player or the same game while totally ignoring the similarity of the other end. This is undesirable. In contrast, attributed random walk measures such similarity by attributes and allows to transit to similar nodes even if they are not connected.

%that are similar to the unshared node.

%For example, the two edges may have the same player but very dissimilar games. Although both games are played by the same user, they can be totally different relationships due to user preference. Another example is that two adjacent edges can be associated with the same game. Individual diversity make it very risky to claim that the two relationship are similar

%Given an edge, a simple topology-based walk iteratively visits consecutive edges that share the same end and totally ignores how similar the unshared end is. 

%For a relationship (i.e., an edge) to be similar, it is not sufficient to have only one end to be similar. However, when choosing the next node to walk by, a topology-based method does not take into consideration the attribute similarities between the target nodes and previous node. %Thus the edges it provides may be unreasonable.

% Second, for new new edges with very few neighbors, attributes can be useful information to produce reasonable contexts.

\begin{figure} 
\centering
\includegraphics[width=.33\textwidth, height=.3\textwidth]{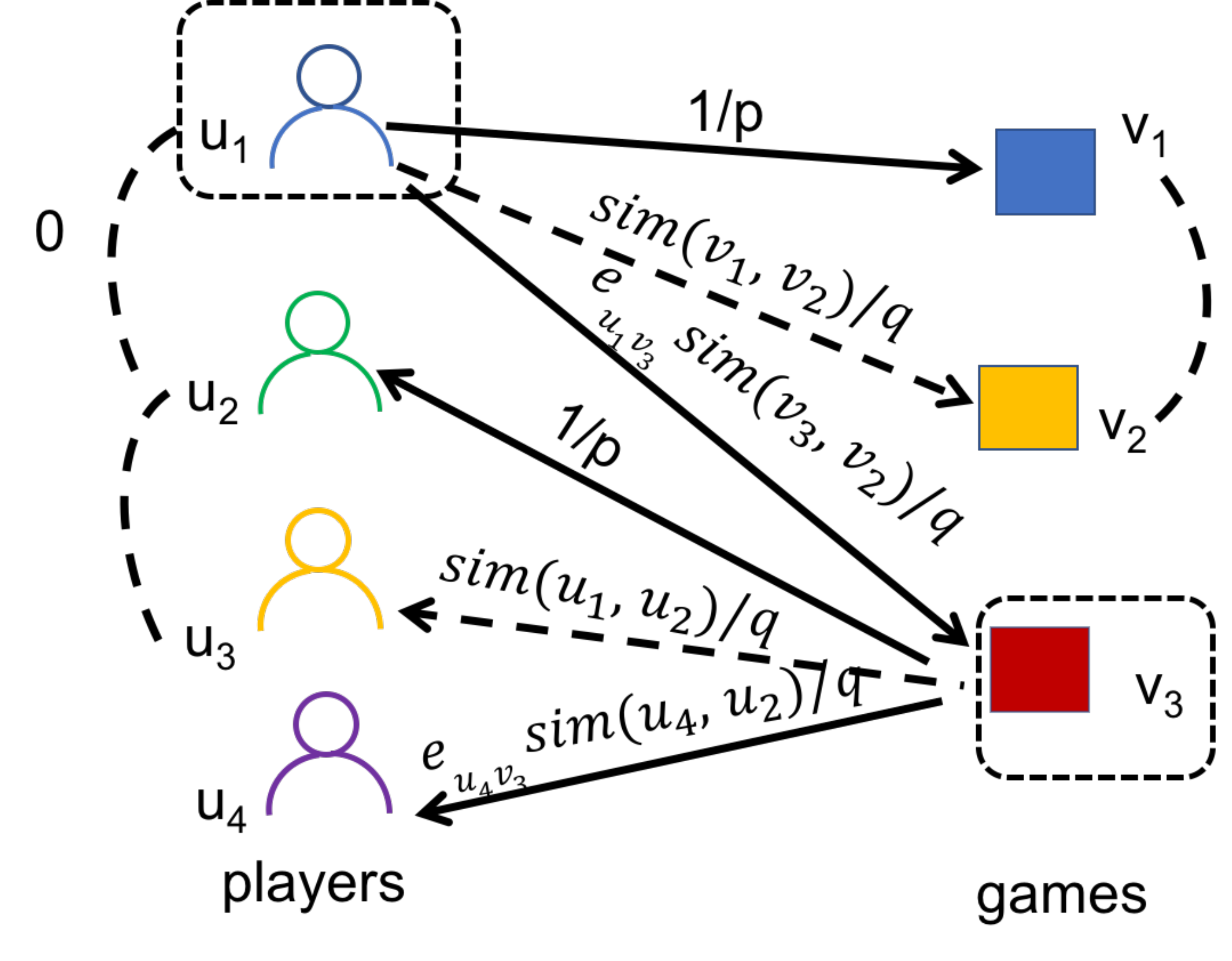} % Give the name of the image here.
\vspace{-1mm}
\caption{An illustrative example of attributed random walk in an attributed bipartite graph}
\label{fig:attributed_random_walk}
\vspace{-4mm}
\end{figure}

To this end, we propose a novel attributed random walk technique that takes into account both topological adjacency and attribute similarities to make the transition decision of the walk. Fig.~\ref{fig:attributed_random_walk} gives an illustrative example of attributed random walk in an attributed bipartite graph. For clarity, we omit the time index in the following discussion. The \emph{solid line} indicates that there exists an edge in the attributed bipartite graph. We denote the type of node $o$ by $type(o)$. A node's type can be of value either player or game. The \emph{dashed directed} lines do not exist in the original attributed graph but may be considered as transitions by our attributed random walk due to attribute similarities.

It is time-consuming to calculate pairwise similarities between a node and all other nodes with the same type. For this reason, we add augmented edges for a proportion of the same-type nodes. For example, the augmented edges of node $o_{1}$ are $\{(o_{1},o): sim(o_{1}, o) > 1 - \epsilon, type(o_{1}) = type(o)\}$, where $0<\epsilon<1$ is a filtering parameter and:
\begin{align}
    sim(o_{1}, o) := \dfrac{\mathbf{x}_{o_{1}}\cdot \mathbf{x}_{o}}{\|\mathbf{x}_{o_{1}}\| \ \|\mathbf{x}_{o}\|}.
\end{align}
The \emph{dashed undirected} lines in Fig.~\ref{fig:attributed_random_walk} represent the added augmented edges between the nodes and their similar same-typed nodes. 

Consider a random walker that just traversed edge $(v_{1}, u_{1})$ in Fig.~\ref{fig:attributed_random_walk} and now resides at node $u_{1}$. Now the walker needs to decide which node to transit to. Since attribute similarities matter in this walk, the walker cannot just evaluate those nodes that are neighbors of $u_{1}$ as suggested in~\cite{grover2016node2vec}. The walker needs to evaluate all nodes of the same type within the two-hop neighborhood of $v_{1}$. We denote the one-hop same-typed adjacent nodes of $v_{1}$ by $N_{1}(v_{1}) = \{v : d(v_{1},v) = 1, type(v) = type(v_{1})\}$, where $d(v_{1}, v)$ is the length of shortest path between $v_{1}$ and $v$. For example, $v_{2} \in N_{1}$ (due to the augmented edge between $v_{2}$ and $v_{1}$). Similarly, we denote the set of two-hop same-typed neighbor nodes of $v_1$ by $N_{2}(v_{1}) = \{v : d(v_{1},v) = 2, type(v) = type(v_{1})\}$. Then the transition probability in attributed random walk can be calculated as follows:
\begin{equation}
  p(o|u_{1})=\left\{
  \begin{array}{@{}ll@{}}
    0, & \text{if}\ type(o) \neq type(v_{1}) \\
    \frac{1}{p}, & \text{if} \ o = v_{1} \\
    \frac{sim(v_{1}, o)}{q}, & \text{if} \ o\in N_{1}(v_{1})\\
    \frac{sim(v_{1}, o) e_{u_{1}o}}{q}, & \text{if}\ o\in N_{2}(v_{1}) \\
  \end{array}\right.
\end{equation}
where $p$ and $q$ are normalization constants used to control the walk strategy. Recall that $e_{u_{1}o} = 1$ if there is an edge between $u_{1}$ and $o$. The attributed walk in an attributed bipartite graph walks through different types of nodes repeatedly. It enlarges the probability that any two consecutive edges in a path are similar, and thus the probability that the whole set of edges on the attributed random walk path are similar. As an example, suppose we are at $u_{1}$ from $v_{1}$ as shown in Fig. (\ref{fig:attributed_random_walk}). Since $v_{1}$ and $v_{2}$ have high similarities, the walker may transit to $v_{2}$ and produce a path with edges $(v_{1}, u_{1})$ and $(u_{1}, v_{2})$ although $u_{1}$ is not connected to $v_{2}$ in the bipartite graph.

Our proposed method shares some similarities with the idea of~\cite{ahmed2017framework}. However, in their work attribute similarities have no influence on which nodes a walker goes through. In our proposed method, when choosing the next node to visit, there is a nonzero probability to choose those that are not connected to the present node but share similar attributes to the previous node. For those that are connected to the present node, the probability to choose from them is weighted by the similarity between the descendant and the ancestor nodes. In this way, an edge can be the context of another even when they are not adjacent but similar in both ends.

\section{Experimental Evaluation}
\label{section:experimental_evaluation}
In this section, we conduct a comprehensive experimental evaluation over the large-scale real data collected from the Samsung Game Launcher platform. We compare our semi-supervised model (referred to as {\tt SS} in the sequel) with the following state-of-the-art models for mobile game churn prediction:
\begin{itemize}
    \item {\tt LR}: the logistic regression based solution used in~\cite{kim2017churn,xie2016predicting, xie2015predicting}
    \item {\tt RS}: the supervised variant of our model, in which the loss function contains only the supervised component and the regularization term
    \item {\tt DT}: a decision tree based solution
    \item {\tt RF}: the random forests based solution used in~\cite{kim2017churn,xie2015predicting}
    \item {\tt SVM}: the SVM based solution used in~\cite{xie2016predicting,xie2015predicting}.
\end{itemize}

In the experiments, we consider the churn duration $T=14$, but again the proposed solution is not restricted to any particular choice of $T$.

\subsection{Dataset and Feature/Label Construction}
%\noindent\textbf{Dataset.} We evaluate the performance of our proposed model and the state-of-the-art competitors on real industry data from the mobile gaming platform
%Samsung Game Launcher platform. 
%This dataset contains 3.6 
%million records, consisting of 15,000 players and 20,000 games, collected in a 4-months period.

% A basic description of our data
Two anonymous datasets were collected \emph{independently} from the Samsung Game Launcher platform within a 4-months period (from August 1st, 2017 to November 30th, 2017) with users' consent, one from the users in \emph{USA} and the other from the users in \emph{Korea}. We summarize the key statistics of these two datasets in Table~\ref{table:data}.

\begin{table}[t]
\centering
\caption{Dataset statistics}
\begin{tabular}{|c|c|c|c|}
\hline
\textbf{Dataset}   & \textbf{\# of users}   & \textbf{\# of games} & \textbf{\# of play records} \\\hline
\emph{USA}       & 15,000      & 19,705      & 76,468,301 \\ \hline
\emph{Korea}    & 25,000      & 18,470      & 106,544,313   \\ \hline
\end{tabular}
\vspace{-0.3cm}
\label{table:data}
\end{table}

The collected data contains three major types of information: (1) play history, (2) game profiles, and (3) user information. Each play record in the play history contains the anonymous user id, the game package name, and the timestamp of play. It is also accompanied with rich contextual information, such as WiFi connection status, screen brightness, audio volume, etc. Game profiles are collected from different game stores, which include features like genre, developer, number of downloads, rating values, number of ratings, etc. User information contains the device model, region, OS version, etc. However, data within games (e.g., levels) is not available due to privacy concerns.

\begin{figure}[t] 
\centering
\includegraphics[width=.45\textwidth]{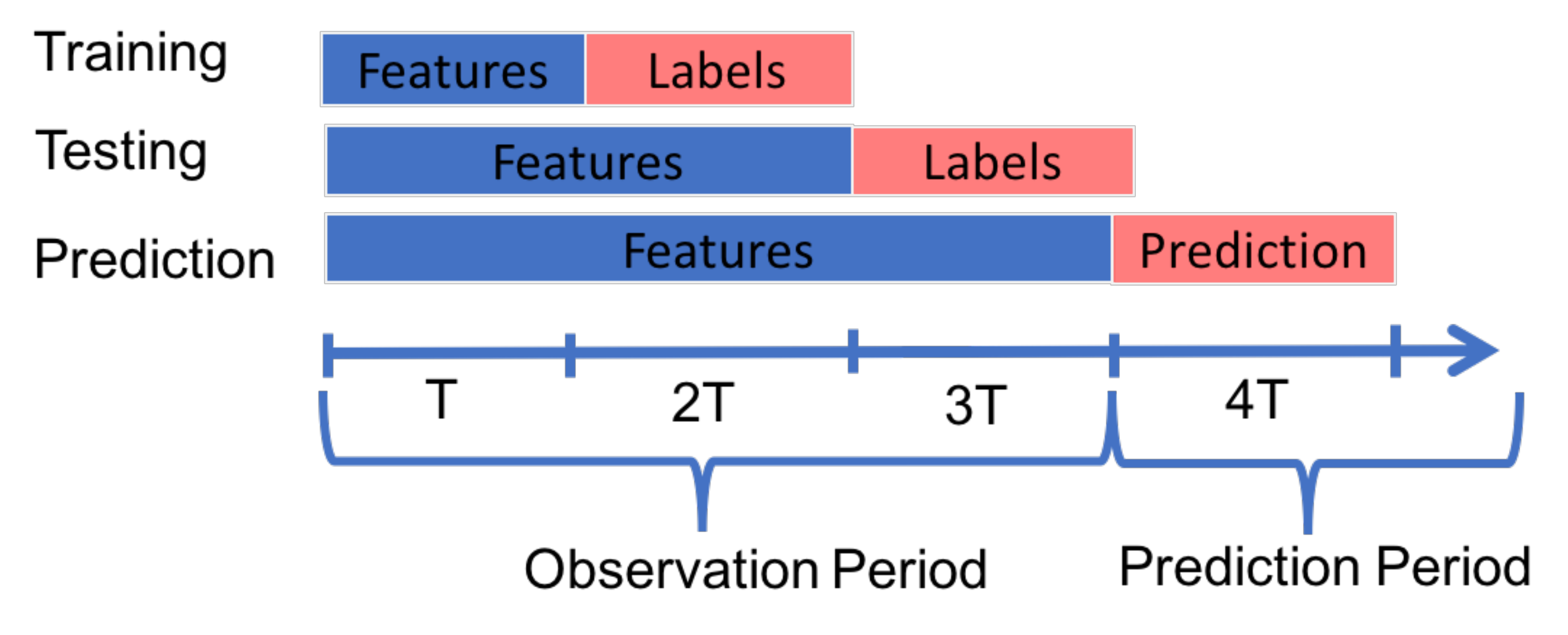}
\vspace{-4mm}
\caption{Feature and label generation process}
\label{fig:train_period}
%\vspace{-2mm}
\end{figure}

Features and labels need to be generated with care in order to avoid data leakage. The generation process is illustrated in Fig.~\ref{fig:train_period}. Labels and features are taken from disjoint periods to avoid data leakage. The model is trained based on features and labels within the observation period. Labels are whether a player churns a game on that day. Features are constructed from historical data before the day to be predicted. The training set and testing set are split by label days in chronological order~\cite{chamberlain2017customer}, for example, taking the labeled data in the first 2/3 of the observation period as the training set and the remaining 1/3 as the testing set. This is also to ensure that there is a time difference between the testing set and the training set.

% \vspace{-2mm}

\subsection{Experimental Settings} 
We tune the hyperparameters, including learning rate, batch size, regularization terms, number of layers and number of neurons per layer,  based on the model performance on the testing datasets. A grid search on those parameters is performed and the combination yielding the best performance is chosen. The regularization parameters $\{\lambda_{i}\}_{i=0}^{4}$ are all set to be 1.  $\alpha$, $\beta$, and $\gamma$ in Equation (\ref{equ:objective}) are chosen to be 0.02, 0.01, 1e-5, respectively. $\epsilon$, $p$, and $q$ discussed in Section~\ref{subsection:attributed_random_walk} are set to be 1, 1, and 0.05, respectively. The parameters used for training the deep neural network are summarized below.

\noindent\textbf{Parameter settings of the \emph{USA} dataset}:
\begin{itemize}
    \item Player feature dimension: 10,042
    \item Game feature dimension: 10,042
    \item Player-game feature dimension: 30
    \item Learning rate: the initial value is 0.017 and decay by $\eta = \eta_0/ (1 + k/2)$, where $k$ is the number of epochs 
    \item Number of neurons: input layers 30, embedding layers 50, output layers 380K
    \item Number of epochs: 6-8
    \item Batch size: 1,024
    \item Context number per user-game pair: 4
    \item Optimizer: Adam method~\cite{kingma2014adam}
    \item Activation function: rectified linear unit (ReLU)
\end{itemize}

\noindent\textbf{Parameter settings of the \emph{Korea} dataset}:
\begin{itemize}
    \item Player feature dimension: 10,042
    \item Game feature dimension: 10,042
    \item Player-game feature dimension: 30
    \item Learning rate: the initial value is 0.019 and decay by $\eta = \eta_0/ (1 + k/2)$, where $k$ is the number of epochs 
    \item Number of neurons: input layers 30, embedding layers 50, output layers 632K
    \item Number of epochs: 8-12
    \item Batch size: 4,096
    \item Context number per user-game pair: 4
    \item Optimizer: Adam method~\cite{kingma2014adam}
    \item Activation function: rectified linear unit (ReLU)
\end{itemize}

%\textbf{Training method.} We tried two methods: 1) co-train, 2) alternative train. The co-train results is  better then alternative train%

\begin{table}[t]
\centering
\caption{Performance of different prediction models on \emph{USA}}
{\small
\begin{tabular}{|p{2cm}|p{1.8cm}|p{1.3cm}|p{1.3cm}|}
\hline
Model & AUC & Recall & Precision
\\ \hline
{\tt SS} & \textbf{\underline{0.82}} &  \textbf{\underline{0.78}} & 0.32
\\ \hline
{\tt RS} & 0.77 & 0.75 & 0.27
\\ \hline
{\tt LR} & 0.66 & 0.38 & 0.26
\\  \hline
{\tt DT} & 0.59 & 0.28 & 0.32
\\ \hline
{\tt RF} & 0.75 & 0.31 & \textbf{\underline{0.41}} 
\\ \hline
{\tt SVM} & 0.61 & 0.78 & 0.18
\\ \hline
\end{tabular}
}
\label{table:performanceusa}
\vspace{-2mm}
\end{table}

\begin{table}[t]
\centering
\caption{Performance of different prediction models on \emph{Korea}}
{\small
\begin{tabular}{|p{2cm}|p{1.8cm}|p{1.3cm}|p{1.3cm}|}
\hline
Model & AUC & Recall & Precision
\\ \hline
{\tt SS} & \textbf{\underline{0.82}} &  \textbf{\underline{0.70}} & 0.34
\\ \hline
{\tt RS} & 0.76 & 0.70 & 0.25
\\ \hline
{\tt LR} & 0.67 & 0.59 & 0.21
\\  \hline
{\tt DT} & 0.58 & 0.26  & 0.30
\\ \hline
{\tt RF} & 0.73 & 0.27 & \textbf{\underline{0.42}} 
\\ \hline
{\tt SVM} & 0.63  &  0.67 & 0.18
\\ \hline
\end{tabular}
}
\label{table:performancekorea}
\vspace{-2mm}
\end{table}

\subsection{Experimental Results}
We use three widely-used evaluation metrics to compare the performance of different models. The most important metric with respect to the business goals is \emph{the area under the ROC curve} (AUC). Following previous studies~\cite{kim2017churn,xie2016predicting, xie2015predicting}, we also consider \emph{precision} and \emph{recall}. Accuracy is not used because our data is imbalanced with around $85\%$ negative instances in the \emph{USA} dataset and $86\%$ negative instances in the \emph{Korea} dataset. 

\begin{figure}[t]
%\centering 
$\begin{array}{c c}
    \multicolumn{1}{l}{\mbox{\bf }} & \multicolumn{1}{l}{\mbox{\bf }} \\
    \hspace{-5.0mm}\scalebox{0.27}{\includegraphics[width=\textwidth]{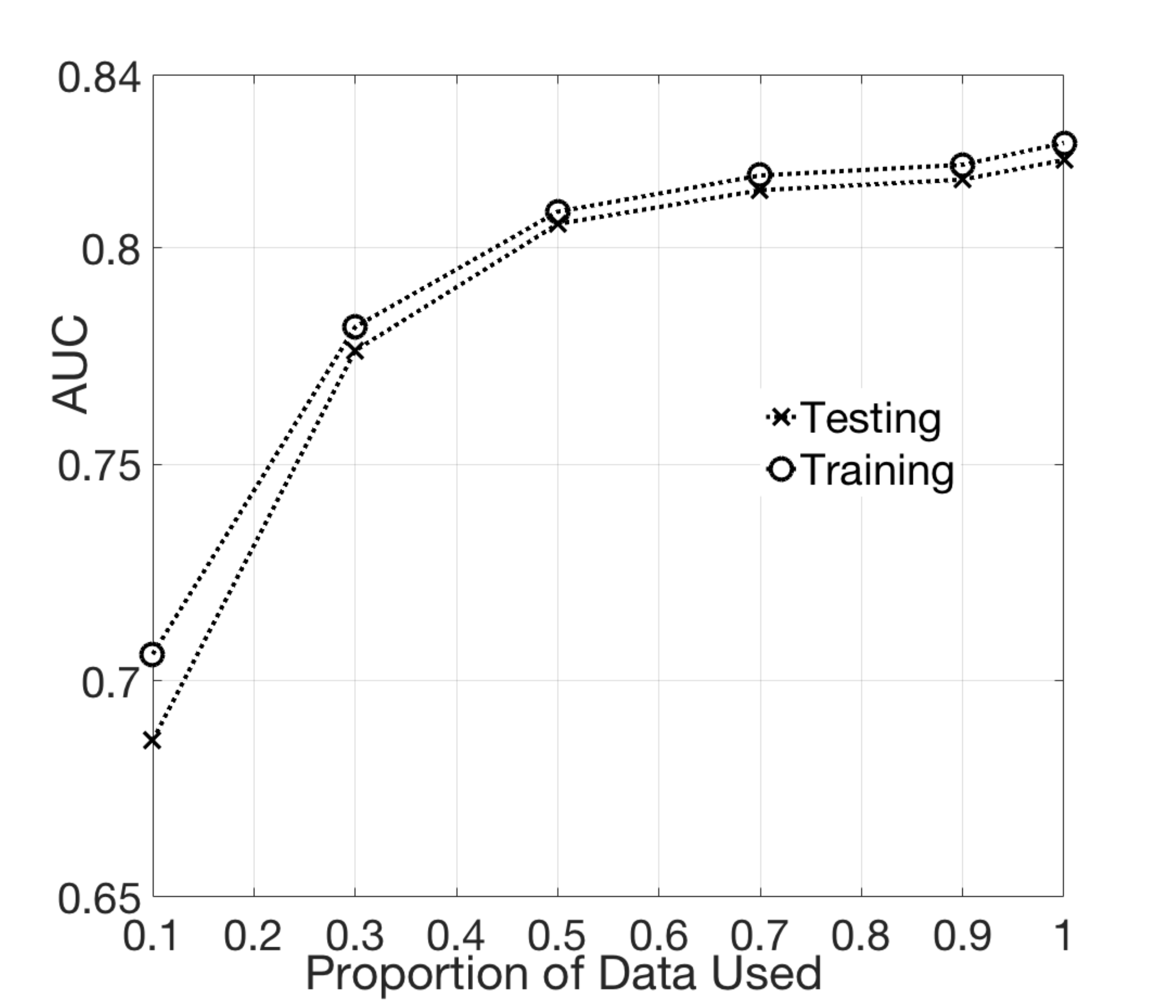}} & \hspace{-6.2mm} \scalebox{0.27}{\includegraphics[width=\textwidth]{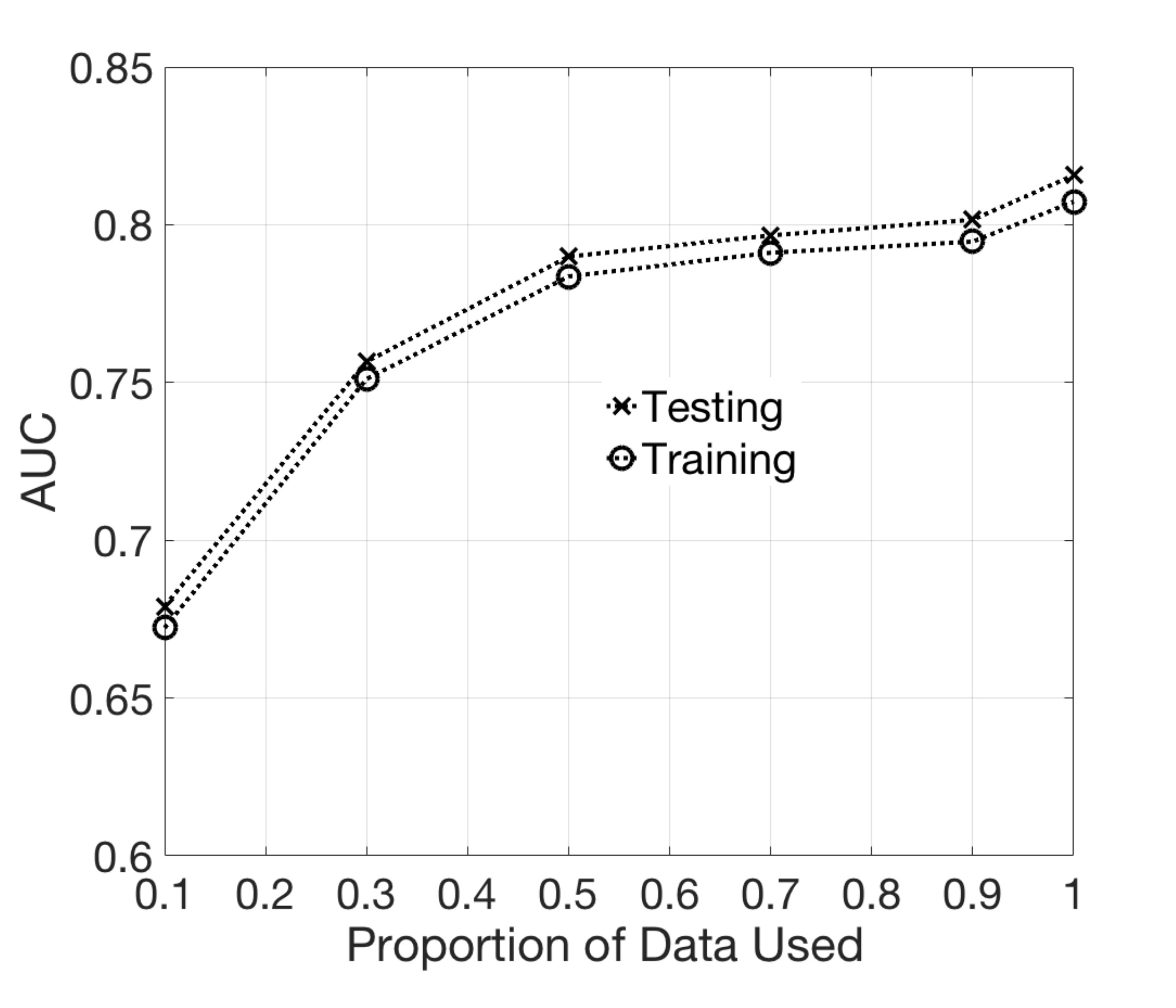}} \\
    \hspace{-2mm}\mbox{\scriptsize (a) AUC on \emph{USA}} & \hspace{-5mm} \mbox{\scriptsize (b) AUC on \emph{Korea}} \\ [-0.0cm]
    \end{array}$
\caption{Comparison of AUC for {\tt SS} in training and testing}
\label{fig:training_testing}
\vspace{-6mm}
\end{figure}

We report the main experimental results in Table~\ref{table:performanceusa} and Table~\ref{table:performancekorea}. It can be observed that in general our model achieves the best AUC and recall on the two testing datasets. Our model outperforms all single models (i.e., {\tt LR}, {\tt DT} and {\tt SVM}) in terms of all the three metrics. In particular, it is worth mentioning that {\tt SVM} achieves a high recall at the cost of a very low precision. This is because it makes a large number of false positives. This fact makes it less useful for business decision making. Compared with the ensemble method {\tt RF}, which has been considered so far the best method in the field, our model still achieves $34\%$ AUC improvement and $250\%$ recall improvement. {\tt RF} achieves the highest precision on the dataset. However, we point out that this number is actually misleading because it can easily overfit and only recognize a small proportion of churn labels. Meanwhile, its recall is poor, making it difficult to meet business requirements of churn prediction (e.g., targeted promotion campaigns). We also carefully compare the AUC of {\tt SS} in training and testing in Fig~\ref{fig:training_testing}. Since the curves are very close, it can be learned that our model is neither overfitting nor underfitting.

% We provide a further analysis based on confusion matrices of {\tt RF} and {\tt SS} in Table~\ref{table:confusion_matrix_random_forest} and Table~\ref{table:confusion_matrix_semisupervised}, respectively. We can observe that {\tt RF} predicts the majority of instances as negative and misses $70\%$ of churn cases. In contrast, our semi-supervised model captures $78\%$ of positive cases by predicting merely $37\%$ of total cases as positive. Recall that one of the business use cases for game churn prediction is targeted marketing campaigns. The performance of {\tt RF} cannot achieve any meaningful targeting capability.

\begin{figure}[t]
%\centering 
$\begin{array}{c c}
    \multicolumn{1}{l}{\mbox{\bf }} & \multicolumn{1}{l}{\mbox{\bf }} \\
    \hspace{-5.0mm}\scalebox{0.27}{\includegraphics[width=\textwidth]{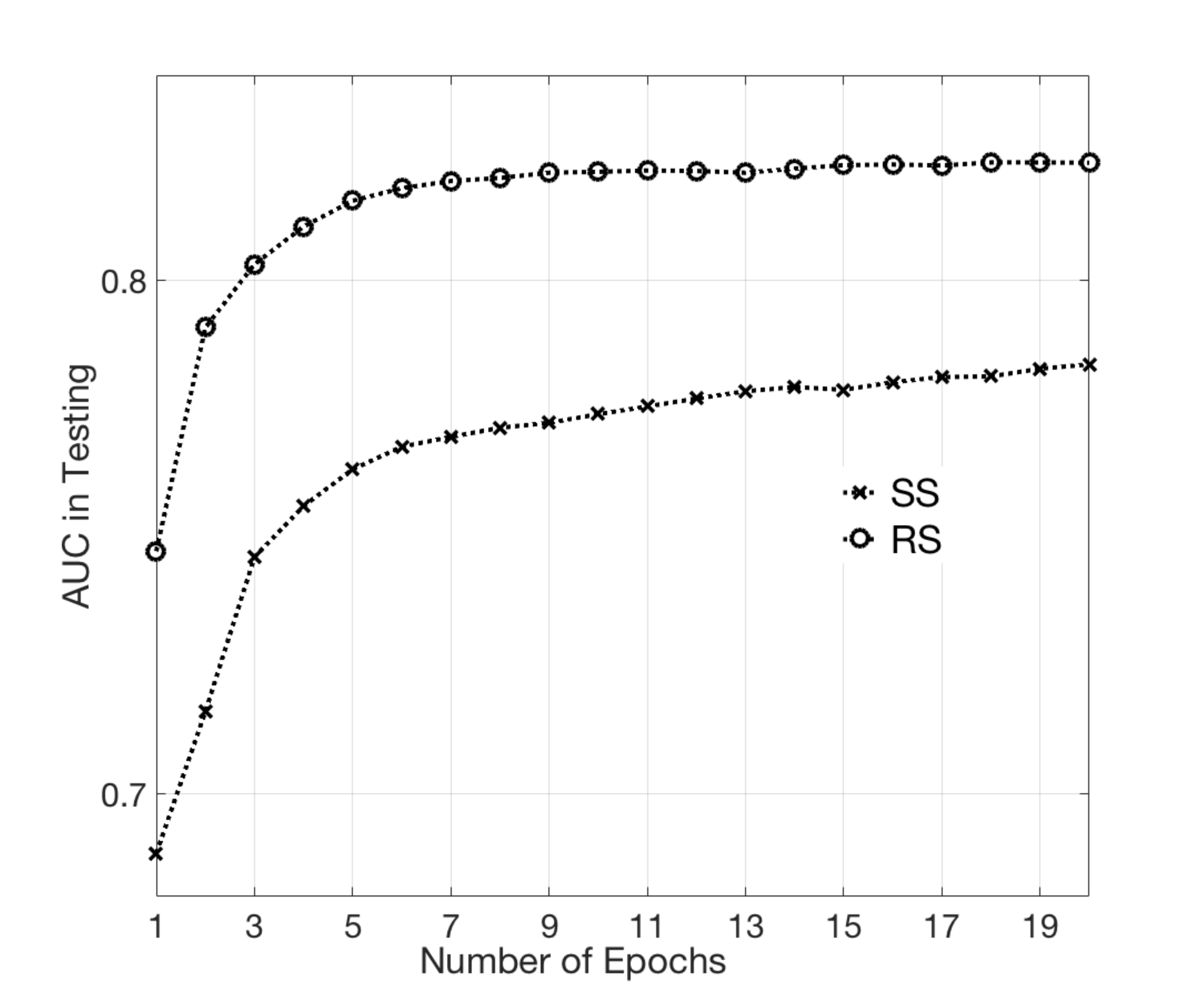}} & \hspace{-5.8mm} \scalebox{0.27}{\includegraphics[width=\textwidth]{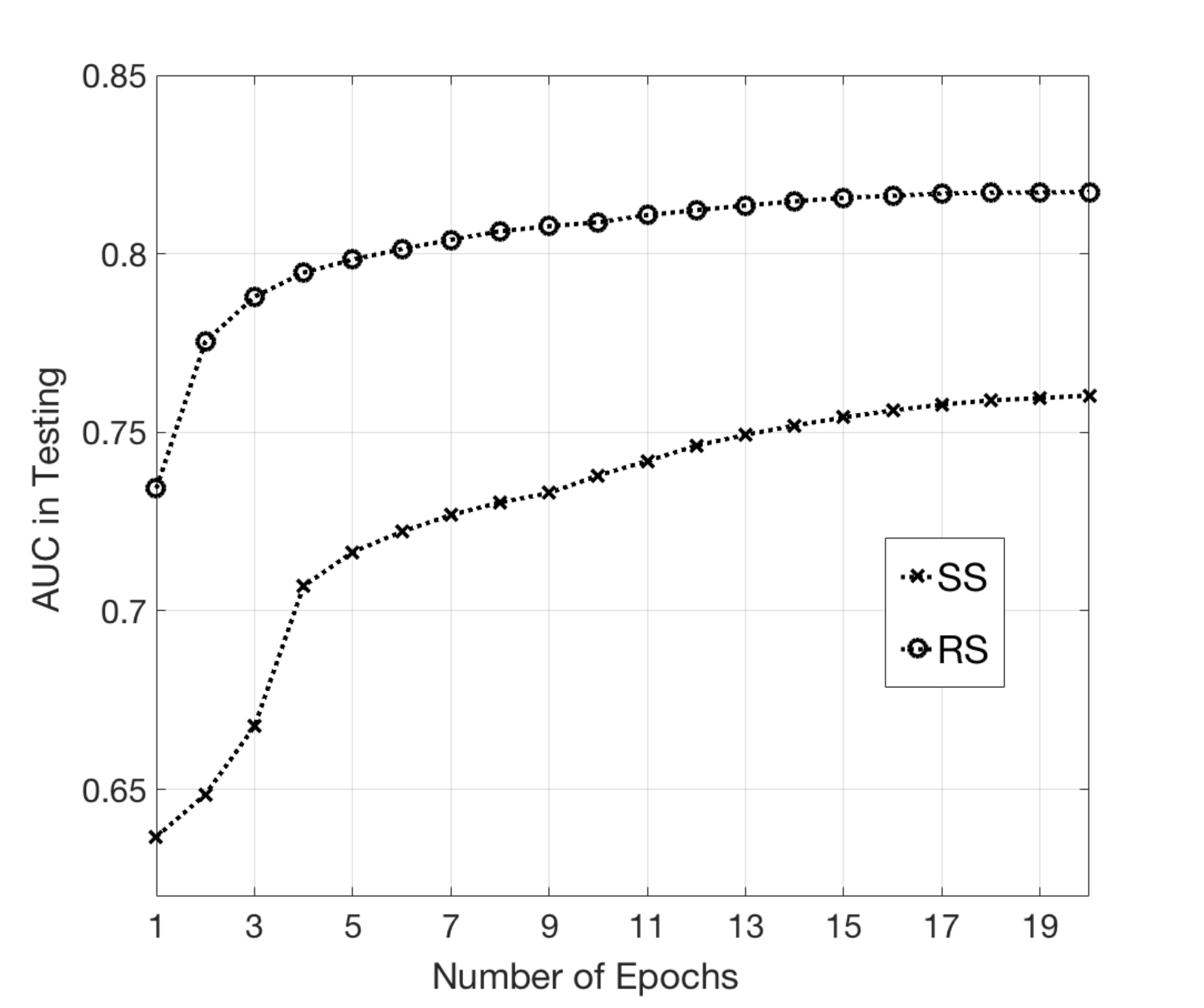}} \\
    \hspace{-2mm}\mbox{\scriptsize (a) AUC on \emph{USA}} & \hspace{-5mm} \mbox{\scriptsize (b) AUC on \emph{Korea}} \\ [-0.0cm]
    \end{array}$
\caption{AUC comparison between {\tt SS} and {\tt RS} under different numbers of epochs}
\label{fig:auc_ss_rs}
\vspace{-4mm}
\end{figure}

The performance difference between {\tt SS} and {\tt RS} justifies the benefits of incorporating unsupervised loss and temporal loss in the objective function. We provide a further comparison between {\tt SS} and {\tt RS} with respect to the number of epochs in Fig.~\ref{fig:auc_ss_rs}. Both models take 5-7 epochs to reach relatively stable performance. We observe that {\tt SS} outperforms {\tt RS} in general under a different number of epochs and for both Korea users and USA users.

%\begin{figure}[htbp]
%\begin{minipage}[t]{0.485\linewidth}
%\centering
%    \includegraphics[width=0.85\linewidth]{figures/auc_epoc.pdf}
%    \caption{AUC comparison between {\tt SS} and {\tt RS} under different numbers of epochs in USA.}
%    \label{figure:case 1-1}
%\end{minipage}%
%    \hfill%
%\begin{minipage}[t]{0.485\linewidth}
%\centering
%    \includegraphics[width=0.85\linewidth]{figures/auc_epoc_korea.pdf}
%    \caption{AUC comparison between {\tt SS} and {\tt RS} under different numbers of epochs in Korea.}
%    \label{figure:case 1-2}
%\end{minipage} 
%\end{figure}

%----------------------------------------------------

\begin{figure}[t]
%\centering 
$\begin{array}{c c}
    \multicolumn{1}{l}{\mbox{\bf }} & \multicolumn{1}{l}{\mbox{\bf }} \\
    \hspace{-5.0mm}\scalebox{0.27}{\includegraphics[width=\textwidth]{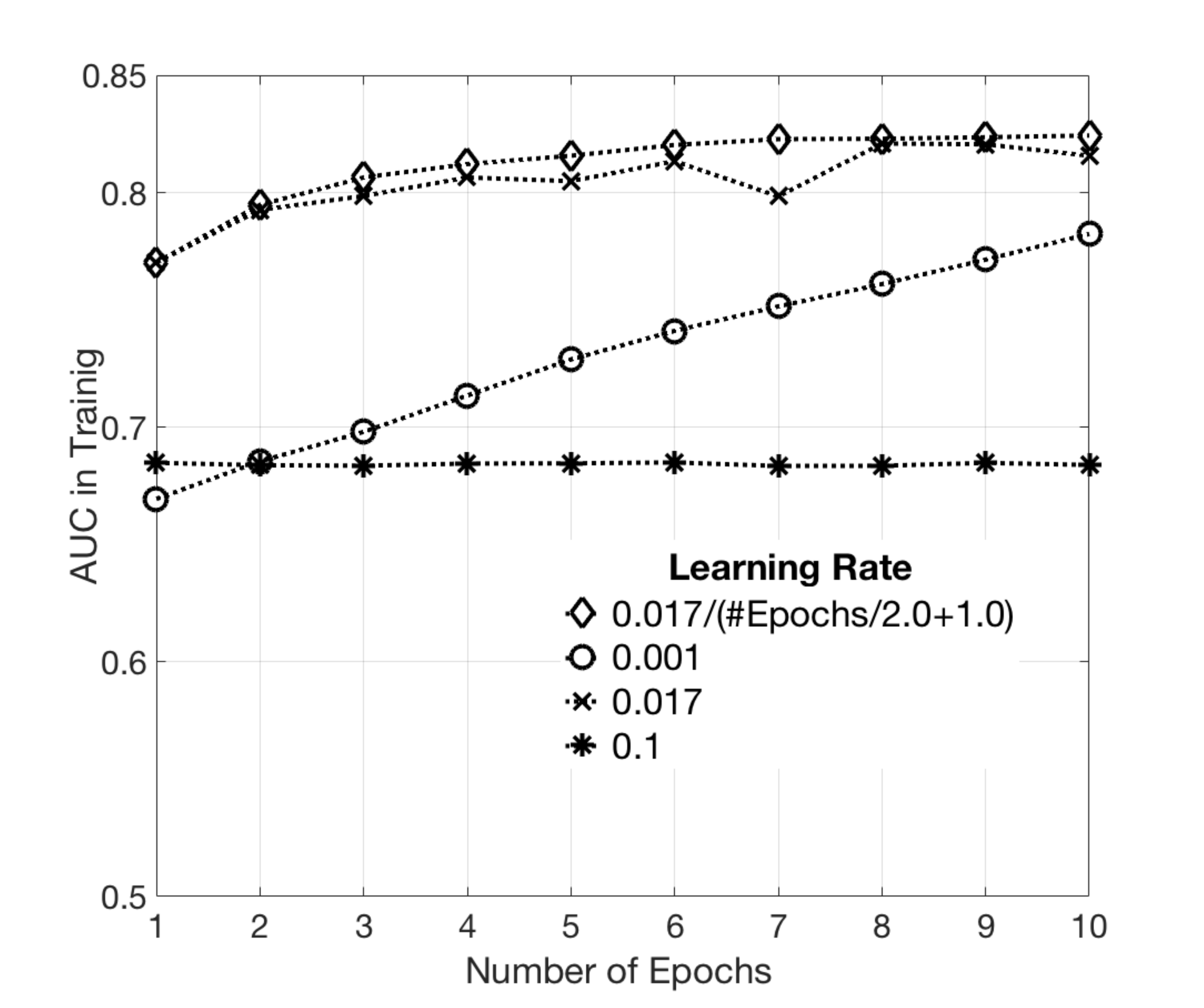}} & \hspace{-5.8mm} \scalebox{0.27}{\includegraphics[width=\textwidth]{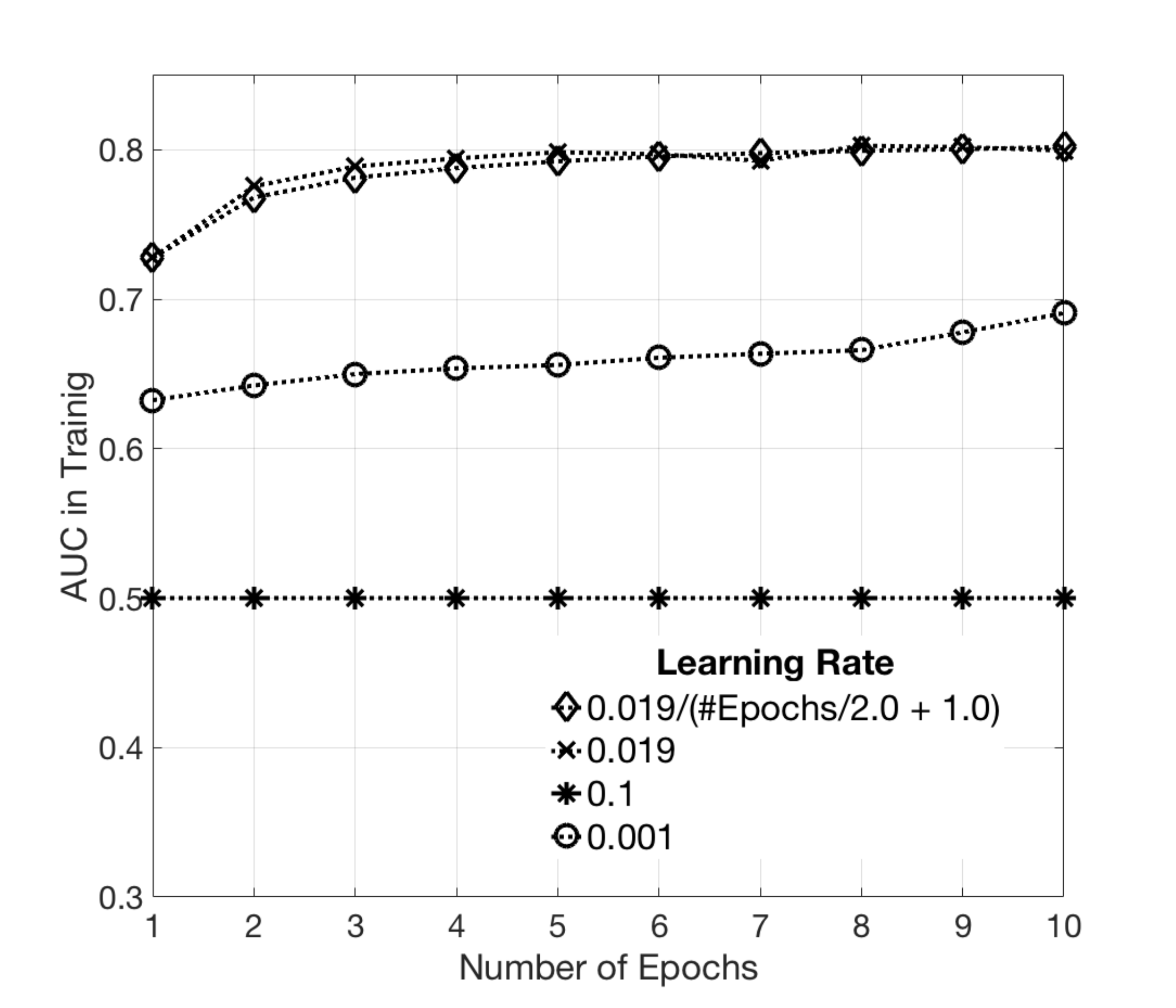}} \\
    \hspace{-2mm}\mbox{\scriptsize (a) AUC on \emph{USA}} & \hspace{-5mm} \mbox{\scriptsize (b) AUC on \emph{Korea}} \\ [-0.0cm]
    \end{array}$
\vspace{-2mm}
\caption{Comparison of AUC for {\tt SS} in training under different learning rates}
\label{fig:learning_rate}
\vspace{-6mm}
\end{figure}

\begin{figure}[t]
%\centering 
$\begin{array}{c c}
    \multicolumn{1}{l}{\mbox{\bf }} & \multicolumn{1}{l}{\mbox{\bf }} \\
    \hspace{-5.0mm}\scalebox{0.27}{\includegraphics[width=\textwidth]{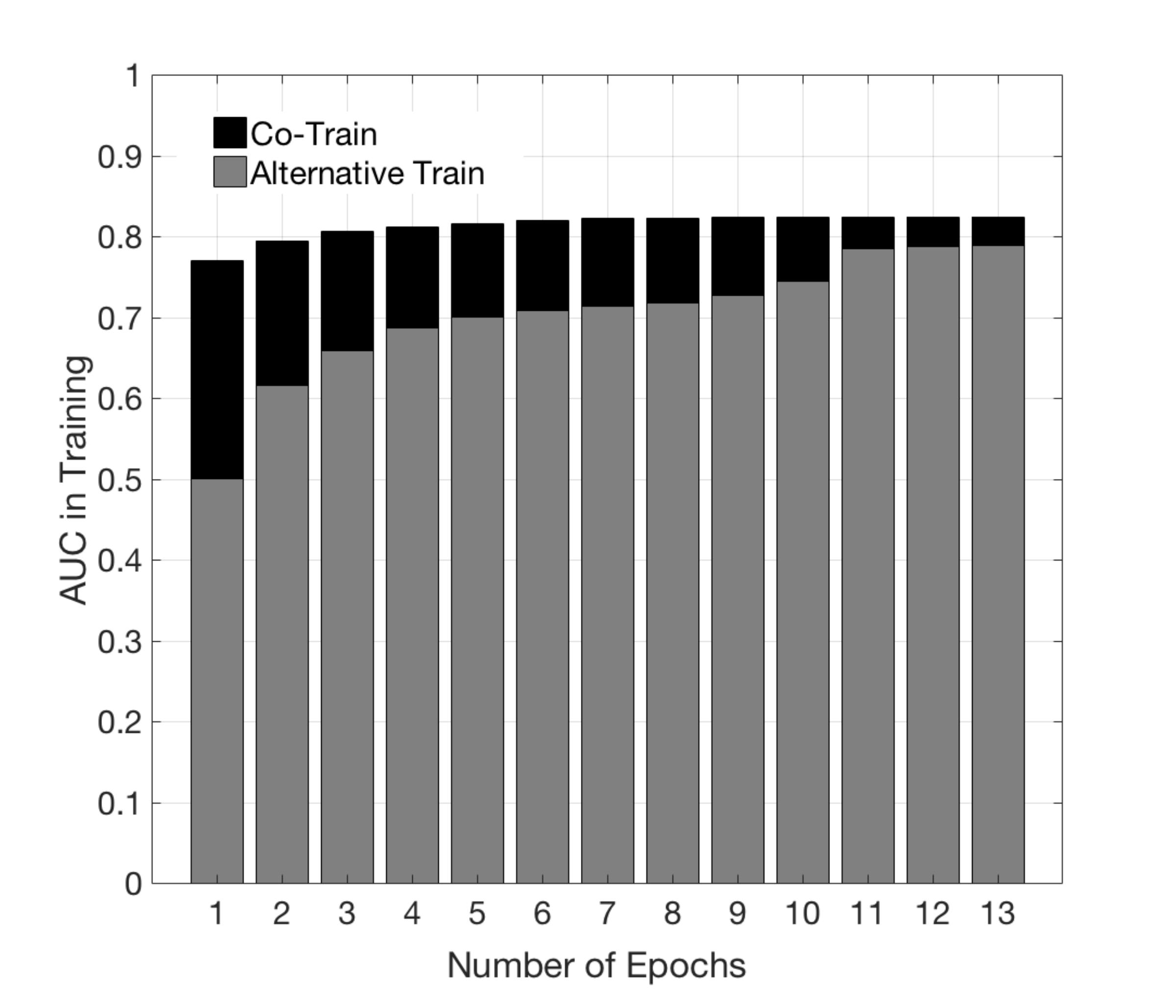}} & \hspace{-5.8mm} \scalebox{0.27}{\includegraphics[width=\textwidth]{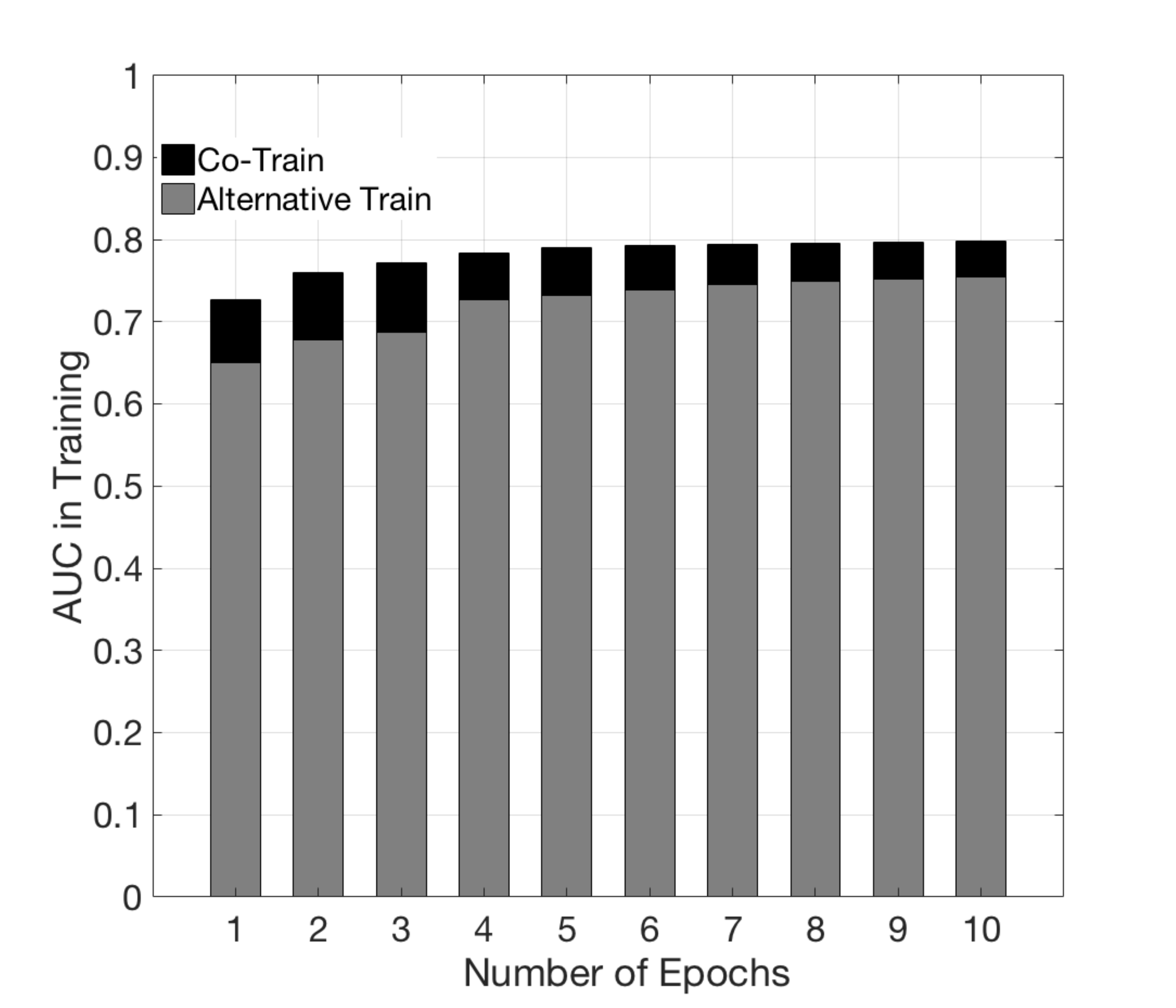}} \\
    \hspace{-2mm}\mbox{\scriptsize (a) AUC on \emph{USA}} & \hspace{-5mm} \mbox{\scriptsize (b) AUC on \emph{Korea}} \\ [-0.0cm]
    \end{array}$
\vspace{-2mm}
\caption{Comparison of AUC for {\tt SS} in training under different training methods}
\label{fig:training_method}
\vspace{-2mm}
\end{figure}

Since the architecture of our DNN is novel and unique (e.g., contain both supervised outputs and unsupervised outputs), we expose more details on how we choose parameters and train the model. A comparison of AUC in training under different learning rates is given in Fig.~\ref{fig:learning_rate}. It can be observed that the choice of the learning rate greatly influences the model performance after the initial epoch. We experimentally find that $0.1$ is too large for the learning rate, which makes the step in gradient descent too large to find a good minimum and that $0.001$ is too small, making it converge very slowly to the optimal point. Therefore we experimentally test learning rates between $0.1$ and $0.001$ and find that $0.017/\text{(\#Epochs/2+1)}$ works best for training on \emph{USA} while $0.019/\text{(\#Epochs/2+1)}$ works best for training on \emph{Korea}.

For the supervised component and unsupervised component, we try two different training methods: \emph{co-train} and \emph{alternative train}. Co-train means that we simultaneously train the supervised loss function and the unsupervised loss function; alternative train means that we alternately train the unsupervised component with the unsupervised loss function and the supervised component with the supervised loss function. This is a widely-used training method for similar structures~\cite{yang2016revisiting, liang2017seano}. It is interesting to observe that in general co-train outperforms alternative train in terms of AUC under a different number of epochs as shown in Fig. \ref{fig:training_method}. Therefore, we choose co-train as the final training methods in our experiments.

%---------------------------------------------------

\section{Related Work}\label{section:related_work}
In this section, we review two categories of existing research that are relevant to this paper. The first category includes the existing works for (game) churn prediction. The early works~\cite{hadiji2014predicting, xie2015predicting, runge2014churn, tamassia2016predicting, xie2016predicting, drachen2016rapid} are based on more traditional machine learning models, such as logistic regression, random forests, SVM, naive Bayes, etc., and are experimentally evaluated on extremely small numbers of games (i.e., less than five). As shown in our experiments, their performance on large-scale real data with tens of thousands of mobile games and hundred of millions of user-app interactions is generally not satisfactory. In addition, we also observe scalability issues when they are applied to large-scale data. Some recent research has started to use more advanced models. \cite{perianez2016churn, bertens2017games, viljanen2017playtime} propose to use survival model for churn prediction, in which churn probability is modeled as a function of playtime. Kim et al.~\cite{kim2017churn} achieve good performance by using convolution neural networks (CNN) and long short-term memory networks (LSTM). There are also several recent deep-learning-based studies~\cite{wangperawong2016churn, umayaparvathi2017automated, martins2017predicting} for non-game churn prediction problems, which report better performance. This motivates us to employ deep neural network models. While being a generic solution, our model is able to accommodate the unique characteristics of mobile gaming. We provide a comparison between all existing works and ours in Table~\ref{table:references}.

\begin{table}[t]
% \def\arraystretch{.8}% 
% \small
\centering
\caption{Comparison between this paper and existing works in data size and key techniques}
{\small
\begin{tabular}{|p{0.7cm}|p{2.7cm}|p{4.3cm}|}
%\begin{tabular}{|c|c|c|}
\hline
\textbf{Paper} & \textbf{Data size} & \textbf{Key techniques} \\ \hline
\cite{hadiji2014predicting} & 5 games \newline 50 thousand users & Decision tree, naive Bayes
\\ \hline
\cite{runge2014churn} & 2 games \newline 10 thousand users & Hidden Markov Model combined with a single layer neural network
\\ \hline
\cite{perianez2016churn, bertens2017games} & 1 game \newline 3 thousand users & Survival ensembles
\\ \hline
\cite{viljanen2017playtime} & 1 game \newline 1 thousand users & Survival model
\\ \hline
\cite{tamassia2016predicting} & 1 game \newline 10 thousand users & Hidden Markov Model
\\ \hline
\cite{xie2015predicting} & 2 games \newline 1 thousand users & SVM, decision tree, logistic regression
\\ \hline
\cite{drachen2016rapid} & 1 game \newline 130 thousand users & Heuristic decision tree
\\ \hline
\cite{xie2016predicting}   & 3 games \newline 60 thousand users & Logistic regression, decision tree, SVM
\\ \hline
\cite{kim2017churn}   & 3 games \newline 200 thousand users & Logistic regression, random forests, CNN, LSTM
\\ \hline
Ours & \textbf{40,000} games \newline 40 thousand users & Deep attributed edge embedding
\\ \hline
\end{tabular}
}
\label{table:references}
\vspace{-4mm}
\end{table}

The second category contains recent works on graph embedding. Graph embedding automates the entire process of feature engineering by casting feature extraction as a representation learning problem. It frees models from human bottleneck introduced by handcrafted features and is able to utilize the full richness of data. However, most existing works such as Node2Vec~\cite{grover2016node2vec}, Deep Walk~\cite{perozzi2014deepwalk} and LINE~\cite{tang2015line} are \emph{node-centric} embedding. When it comes to game churn prediction, the entities to be embedded are edges that represent the relationship between players and games. Very limited work has studied edge embedding. Abu-El-Haija et al.~\cite{abu2017learning} propose to model edge embedding as a function of node embedding, in which the two ends of an edge are first embedded and then passed into a deep neural network with edge embedding as output. This method embeds edges in an indirect way and does not take into account any attribute information. In addition, these models~\cite{grover2016node2vec, perozzi2014deepwalk, tang2015line} all belong to the transductive framework, in which embedding cannot be generated if an object has never appeared in training. However, in our problem new users and new games appear continuously; new relationships between existing users and games may form anytime in the future. Therefore, for game churn prediction the capability of handling new edges is indispensable. 

Several very recent works have proposed the idea of inductive graph embedding~\cite{liang2017seano, hamilton2017inductive, yang2016revisiting}, inspired by which we propose our inductive edge embedding model for churn prediction. Our model improves these works in several major ways. First, our embedded features are learned in a \emph{semi-supervised} manner, where the supervised component is for churn prediction while the unsupervised component is for context recovery. Compared to unsupervised methods, embedding features learned in a semi-supervised way have been shown to achieve better performance~\cite{yang2016revisiting}. Second, our model captures graph dynamics by imposing temporal loss to the embeddings of the same edge in consecutive graph snapshots. Unlike all existing works, where embeddings represent structures or attribute information, embeddings in our model are designed to simultaneously capture contexts and graph dynamics. To our best knowledge, our model is the first to achieve such benefits.

\section{Conclusion}\label{section:conclusion}
Churn prediction of mobile games is a vital research and business problem that is backed up by a billion-dollar market. In this paper, we proposed a novel inductive semi-supervised embedding model for large-scale game churn prediction. Our model jointly learns a prediction function and an edge embedding function that can automatically map handcrafted features to latent features. The contexts of an edge are sampled by a novel attributed random walk technique, in which both topological adjacency and attribute similarities are considered. We modeled the prediction function and the embedding function by deep neural networks, where the embedding component is designed to capture both contextual information and relationship dynamics. We compared our model with several state-of-the-art baseline methods on large-scale real-world data, which is collected from the Samsung Game Launcher platform. The experimental results clearly demonstrate the effectiveness of the proposed model. 

Although the paper focuses on mobile game churn prediction, the proposed method is not restricted to this specific problem and also works for more general problems. Since the inputs of our model only contain the attributes of objects and their relationships at different timestamps, the proposed model can be generalized to fulfill any churn prediction task where the underlying data can be modeled in a similar way, for instance, customer disengagement prediction in membership business (e.g., Apple Music, Costco, and insurance companies) and interest group unsubscription prediction in social networks (e.g., Facebook, Meetup, and Google+).

\bibliographystyle{IEEEtran}
\bibliography{reference}
\end{document}